\documentclass{article}
\usepackage{arxiv}

\usepackage{hyperref}
\usepackage{bm}
\usepackage{cite}
\usepackage{graphics,fancyhdr,graphicx,subfigure}
\usepackage{subfigure}
\usepackage{amsthm,amsmath,amsfonts,latexsym,amssymb}
\usepackage{lineno}
\usepackage{diagbox}
\usepackage{algorithm}
\PassOptionsToPackage{noend}{algpseudocode}
\usepackage{algpseudocode}
\usepackage{multirow,booktabs}
\usepackage{listings}
\usepackage[table]{xcolor}
\usepackage{lscape}
\usepackage{comment}
\usepackage{tabularx}
\usepackage{graphicx}
\usepackage{tabularx}
\usepackage{arydshln}
\usepackage{nomencl}
\usepackage{array}
\usepackage{caption}
\usepackage{breqn}
\usepackage{hyperref} 
\usepackage{lineno}
\usepackage{mathtools}
\usepackage{subfigure}
\usepackage{caption}
\usepackage{lmodern}
\usepackage{calligra}
\usepackage{xspace}
\usepackage[utf8]{inputenc}
\usepackage{url} 
\usepackage{enumerate}
\usepackage[english]{babel}
\newtheorem*{remark}{Remark}

\def\equationautorefname~#1\null{%
  Eq.~(#1)\null
  }
\def\subfigureautorefname~#1\null{%
  Fig.~#1\null
}

\definecolor{listinggray}{gray}{0.9}
\definecolor{lbcolor}{rgb}{0.9,0.9,0.9}
\definecolor{Darkgreen}{RGB}{0,100,0}

\title{Geometry adaptive waveformer for cardio-vascular modeling}

\author{ \hspace{1mm}Navaneeth~N.\\
	Department of Applied Mechanics,\\
	Indian Institute of Technology (IIT) Delhi,\\
	Hauz Khas - 110 016, New Delhi, India \\
	\texttt{navaneeth.n@am.iitd.ac.in} \\
	\And
	\hspace{1mm}Souvik~Chakraborty\\
    Department of Applied Mechanics,\\
    Yardi School of Artificial Intelligence, \\
    Indian Institute of Technology (IIT) Delhi,\\
    Hauz Khas - 110 016, New Delhi, India\\
    University of Illinois Urbana-Champaign (UIUC)\thanks{Fulbright-Nehru Fellow} \\
    \texttt{souvik@am.iitd.ac.in} \\
}

\renewcommand{\shorttitle}{Geometry adaptive waveformer for cardio-vascular modelling}

\hypersetup{
pdftitle={Geometry adaptive waveformer for cardio-vascular modelling},
pdfsubject={q-bio.NC, q-bio.QM},
pdfauthor={Navaneeth N., Souvik Chakraborty},
pdfkeywords={Graph neural operator, Waveformer, Transformers, Cardiovascular modeling},
}
\begin{document}
\maketitle

\begin{abstract}

Modeling cardiovascular anatomies poses a significant challenge due to their complex, irregular structures and inherent pathological conditions. Numerical simulations, while accurate, are often computationally expensive, limiting their practicality in clinical settings. Traditional machine learning methods, on the other hand, often struggle with some major hurdles, including high dimensionality of the inputs, inability to effectively work with irregular grids, and preserving the time dependencies of responses in dynamic problems. In response to these challenges, we propose a geometry adaptive waveformer model to predict blood flow dynamics in the cardiovascular system. The framework is primarily composed of three components: a geometry encoder, a geometry decoder, and a waveformer. The encoder transforms input defined on the irregular domain to a regular domain using a graph operator-based network and signed distance functions. The waveformer operates on the transformed field on the irregular grid. Finally, the decoder reverses this process, transforming the output from the regular grid back to the physical space. We evaluate the efficacy of the approach on different sets of cardiovascular data.

\end{abstract}


\section{Introduction}\label{sec:intro}
Determining the health conditions of patients is a pervasive challenge in modern clinical practice, particularly as the healthcare industry strives to provide more personalized, patient-centered care. Cardiovascular disease (CVD) \cite{gaziano2006cardiovascular} is a prominent example of this, having emerged as one of the leading causes of death globally, contributing to a significant portion of healthcare morbidity and mortality statistics. CVD encompasses a wide range of conditions, from coronary artery disease and hypertension to heart failure and stroke. The ability to accurately assess a patient’s cardiovascular health is paramount for both preventing adverse events and optimizing treatment pathways. In this regards, cardiovascular modeling, which combines advanced imaging techniques with computational simulations, is increasingly being recognized for its clinical utility. By simulating the behavior of the cardiovascular system under various physiological and pathological conditions, these models provide insights that are difficult to obtain through imaging alone. They allow for the prediction of disease progression, the evaluation of surgical or therapeutic interventions, and the personalization of treatment plans. This approach not only enhances the precision of diagnoses but also enables a more strategic and personalized management of cardiovascular diseases, potentially reducing the risk of mortality and improving patient outcomes.

Mathematical modeling and numerical methods have impacted several scientific fields, including the medical domain. Among popular methods, Computational Fluid Dynamics (CFD) has become a prominent approach and is extensively used in simulation-based studies of the cardiovascular system \cite{bao2014usnctam,schwarz2023beyond}. Notably, most of the 3D blood flow simulations utilise CFD \cite{zhong2018application} to obtain patient-specific simulation and further use finite element methods (FEM) \cite{stuhne2004finite} to yield a precise estimation of wall shear stress and velocity. However, the effectiveness of CFD in these models is limited by resolution constraints and the inability to account for the movements of elastic artery walls during blood flow computation \cite{eslami2020effect}. To address the challenge, ensuing research was carried out by considering deformable structures and their interactions with fluids, which resulted in the development of fluid-structure interaction (FSI) techniques \cite{syed2023modeling,crosetto2011fluid}. FSI outperform traditional CFD-based cardio-vascular simulations, offering more realistic flow simulations and deeper insights into blood flow dynamics coupling methods like FEM and finite volume method (FVM). There are also concurrent approaches such as the immersed boundary method (IBM) \cite{tay2011towards,sarkar2017immersed} and lattice Boltzmann methods (LBM) \cite{pontrelli2014lattice}, which are mesh-free approaches and provide grid-independent simulations. Despite the advantages, all the above methods employ CFD simulations in a 3D space, which, in general, are computationally prohibitive in nature, limiting their direct application in the clinical domain.

Reduced-order models (ROMs) \cite{dal2020reduced,zhang2020personalized} have been proposed to mitigate the limitations associated with the above-mentioned high-cost simulations employed for cardiovascular systems. Fundamentally, ROMs for cardiovascular simulations attempts to reduce the number of independent variables based on simplified formulations and assumptions. Subsequently, the output variables are described in a space with reduced independent variables. Among the reduced-order models, the simplest one is a zero-dimensional model \cite{gul2016mathematical, liang2005closed,li2019method}, which is also known as the lumped parameter model. A zero-dimensional model can be devised analogous to an electric circuit \cite{bandola2016identification} wherein the blood flow rate due to the pressure difference is treated as the current flow due to the electric potential difference. The notable limitation of the lumped parameter models is due to the inherent assumption that field variables are independent of spatial variables. Thus, they do not preserve the parameter variation over different branches, resulting in an inadequate representation of the pathological conditions.
In contrast, one-dimensional reduced-order models provide a better representation of the vascular system. These 1-D models \cite{formaggia2003one,soudah2014reduced} are obtained by approximating the three-dimensional Navier-Stokes equations to a single spatial variable by integrating over the cross-sectional area of the vessels. They represent arterial branches as centerline segments, with pressure, flow rate, and vessel wall displacement varying along the axial distance. While the 1-D reduced order model accounts for flow interaction with elastic vessel walls better than lumped-parameter models, they often perform poorly due to the enforced pressure conservation between inlets and outlets without considering critical geometrical parameters like branch angles. While ROM perform well at a fraction of the computational cost of full three-dimensional simulations, they also lead to a considerable computational cost when analysis necessitates repeated simulations. 

Data-driven methods \cite{branen2021data,ye2024data} have emerged as promising alternatives to address the limitations of traditional numerical methods, including reduced-order modeling approaches. Notably, there have been significant advancements \cite{siena2023data,pfaller2024reduced} in data-driven cardiovascular modeling. These methods offer the advantage of adaptability to specific geometries through enhanced interpolation schemes. A parallel approach to data-driven learning methods is proposed to accelerate computational processes by incorporating physical laws into modeling. The method is referred to as physics-informed learning (PIL), and one of the most popular algorithms in this paradigm is the Physics Informed Neural Networks (PINNs). Recent efforts have extended the application of PINNs in cardio systems and vascular modeling \cite{sahli2020physics,alzhanov2024three} as well. Nevertheless, PINNs are limited by their inflexibility in dealing with high dimensional systems and learning solution corresponding to different input conditions require retraining from scratch. A recent approach to circumvent the bottlenecks of PINNs and vanilla deep learning methods is operator learning. Operator learning offers a paradigm shift by learning the solution of the physical system through a function-to-function mapping between inputs and outputs. Several operator learning frameworks have been proposed, built on the universal approximation theorem for operators, including Graph Neural Operator (GNO) \cite{anandkumar2020neural,li2020multipole}, Laplace neural operator \cite{cao2024laplace}, Fourier Neural Operator (FNO) \cite{li2020fourier, kovachki2021universal}, and Wavelet Neural Operator (WNO) \cite{tripura2022wavelet, navaneeth2023physics}. Most of the aforementioned operators are kernel-based operators that learns kernel parameters obtained in the spectral domain. For instance, FNO obtains the spectral space through a fast Fourier transformation, whereas WNO uses a wavelet-transformed space utilizing space frequency localized basis functions \cite{zhang2019wavelet}. These approaches have found applications in many fields, including the biomedical field. One such concurrent work explores the scope of the operator for the detection and quantification of the tumour \cite{tripura2023wavelet}.

While most of the operator learning paradigms excel at learning solution operators, they often lag behind the tasks that involve learning dynamical systems, primarily due to the absence of mechanisms for preserving long-term temporal dependencies. Recent work proposes a new framework to address this limitation by effectively combining elements of wavelet transformation, kernel integral operator, and transformer. The framework, referred to as waveformer \cite{navaneeth2024waveformer}, exploits the strength of the transformer in capturing dependencies between elements within a sequence through a specialized feature known as the attention mechanism \cite{wiegreffe2019attention} and facilitates a continuous kernel integration within the wavelet domain. The waveformer is effectively operated on a regular grid; however, it is not well suited for irregular grids. The geometry of practical systems, such as cardiovascular systems, often has complex geometries that can only be described on irregular grids, posing challenges for the direct application of the waveformer. To address this challenge, we introduce a geometry-adaptable waveformer designed for modeling cardiovascular systems. This approach involves applying specific geometry transformations to map the irregular grid onto a regular grid, allowing the waveformer to operate within this regular latent space. Subsequently, an inverse geometry transformation maps the output of the waveformer back to physical space. The geometry transformations are developed by leveraging recent studies \cite{li2024geometry} that have explored geometry-informed operator approaches. We also propose an efficient version of the geometry adaptive waveformer that learns the solution operator by projecting it into a low-dimensional space. In summary, here, we strive to harness the benefits of waveformer, which can model the dynamical system defined on any general geometry. Finally, we evaluate the performance of the geometry adaptive waveformer through its application to a set of cardiovascular geometries. The results demonstrate the effectiveness of the proposed framework in accurately representing complex anatomical structures.

The remainder of the paper is organized as follows. A technical context of the problem is described in Section \ref{sec:Problem statement}. Details of methedology are elucidated in the Section \ref{sec: Methedology}. Thereafter, the results of the model employed for cardiovascular geometries are provided in the section\ref{sec: Numerical example}. Lastly, the concluding remarks are provided in Section \ref{sec:conclusion}.

\section{Technical Context}
\begin{figure}[t!]
    \centering
    \includegraphics[width=\textwidth]{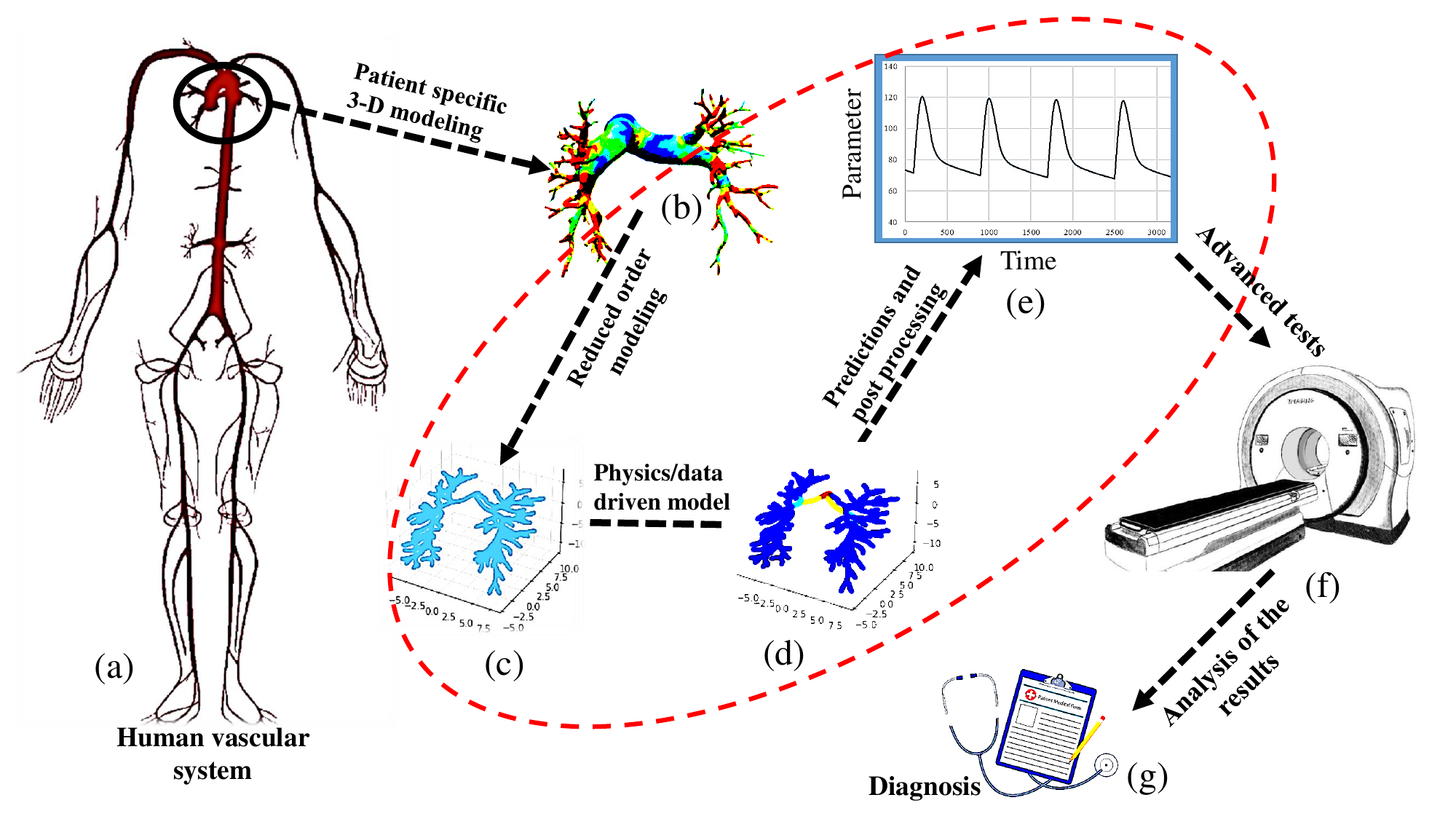}
    \caption{Schematic representing a patient-specific cardiovascular analysis involving computational modeling, post-processing, inference, and diagnosis}
    \label{fig:Shematic_pbstatement}
\end{figure}

This work aims to streamline the workflow of patient-specific cardiovascular analysis, alleviating the computational demands of simulations. Generally, patient-specific cardiovascular analysis and diagnosis involves several steps, starting from data acquisition to diagnosis. Following data acquisition, the process involves segmentation to delineate and understand the different branches of the vascular structure. Subsequently, computational models are developed to simulate the vascular system. After obtaining simulation results, post-processing and analysis are performed to interpret the data. Finally, advanced testing is conducted to arrive at a definitive diagnosis. In this context, we specifically aim to streamline the third step: the development of patient-specific modeling of the vascular system. A schematic illustrating this process is shown in Fig. \ref{fig:Shematic_pbstatement}.

To achieve this, we divide this step into further substeps:
\begin{itemize}
\item 3D Patient-Specific Simulations: Developing detailed computational models of specific cardiovascular segments using 3D simulations.
\item Reduced Order Modeling: Implementing a reduced-order model to simplify the full-scale models, thereby enabling faster and more efficient simulations.
\item Enhanced Data-Driven Modeling: Employing an enhanced data-driven model for training the ROMs, which further improves efficiency by eliminating the need for repeated simulation.
\end{itemize}
For enhanced data-driven modeling, we leverage a waveformer and configure it as a geometry adaptive model, ensuring it can seamlessly handle the complexities of intricate vascular anatomies.

\subsection{Problem statement}\label{sec:Problem statement}
The basic premise of a geometry-adaptive waveformer is to transform an irregular physical space $D_s$  into a regular computational space $D_r$, facilitating efficient operator learning in this transformed domain. Let $\bm s \in \mathcal{S}$ denote the input function, where $\mathcal{S}$ is a Banach space comprising real-valued functions defined over $D_s \subset \mathbb{R}^d$. We define a regular domain $D_r$ such that $D_s \subset D_r \in \mathbb{R}^d$, ensuring that the input functions can be effectively sampled or approximated on this regular grid. To map the input function to a regular domain, we define the set of distance functions $\bm T \subset \mathcal{S}$ such that, for each function, $\bm T \in \mathcal{T}$, has a zero set $Z_T$ that forms a smooth, closed, simply connected sub-manifold. Furthermore, we assume the existence of $\epsilon > 0$ such that: $B_{\epsilon}(x) \cap \partial D = \emptyset$ for every $x \in Z_T$ and $\bm T \in \mathcal{T}$. The sub-manifold  $Z_T$  encloses a volume  $\Omega_T$ within  $D_r$, and we define the remaining domain $D_s$  as ${D_r} \setminus \overline {\Omega}_T$.

Now we define an operator $\mathcal{M}$ that enables a mapping from input function $\bm s$ along with distance function and boundary conditions to output function $\bm u$ such that:
\begin{equation}
     \bm u = \mathcal{M}({\bm s,\bm T,\bm g}),
\end{equation}
where the boundary condition is given by:
\begin{equation}
   \bm u = \bm g \quad \text{on } \partial {D}_s.
\end{equation}
We assume that for each input defined on the geometry configuration $\bm T$, there exists a unique output $\bm u \in \mathcal{U}_{s}$, where $\mathcal{U}_{s}$ is a Banach space. We define an operator $\mathcal{M}_G$, which maps a set of functions comprised of geometry $\bm T$, input $\bm s$, and the boundary condition $\bm g$ to the solution $\bm{u}_r$. Mathematically, the operator can be expressed as: 
\begin{equation}
    \mathcal{M}_{G}: \{\bm s,\bm T,\bm g\} \mapsto \bm{u}_r
\end{equation} 
Further, the extension operator can be defined as a composition of a geometry encoding operator, and $\mathcal{M}_{T}$ operates on regular space $D_r$ such that: 
\begin{equation}
    \mathcal{M}_{G}=\mathcal{M}_{E}\circ \mathcal{M}_{T}. 
\end{equation}
The geometry decoding operator $\mathcal{M}_{D}$ is then applied to the extended solution to obtain the  final output $\bm{u}$  on $D_s$ such that:
\begin{equation}
    \mathcal{M}_{D}:\{\bm{u}_r,\bm{T}\}\mapsto \bm{u}.
\end{equation}
In total the operator $\mathcal{M}$ can be expressed as:
\begin{equation}
    \mathcal{M}=\mathcal{M}_{E} \circ \mathcal{M}_{T} \circ \mathcal{M}_{D}.
\end{equation}
The overarching objective of the work is to devise the operators $\mathcal{M}_{E}$ and $\mathcal{M}_{D}$, which are approximated by graph operators ($\hat{\mathcal{M}}_{E}$ and $\hat{\mathcal{M}}_{D}$), and $\mathcal{M}_{T}$, approximated by waveformer ($\hat{\mathcal{M}}_{T}$) to construct the overall operator $\mathcal{M}$ and employ the framework for modeling cardiovascular systems.

\section{Methodology}\label{sec: Methedology}
The proposed geometry adaptive waveformer essentially aims to model the high dimensional dynamical systems defined on irregular grids. The core concept involves transforming input functions onto a regular latent space and employing the waveformer, which has already proven its effectiveness in learning dynamic systems in regular space. The geometry adaptive waveformer constitutes three components: a graph operator-based geometry encoder, a geometry decoder, and a waveformer. A detailed description of the individual components is discussed here.

\subsection{Graph operator for encoder and decoder}
Before delving into the details of a graph operator, let us consider an operator in a more general setting. Suppose we have a domain $D_a \subset \mathbb{R}^d$ over which we define a parametric input $\bm a \in \mathcal{A}$ and the corresponding output $\bm{u} \in \mathcal{U}$. Given an operator $\mathcal{M}$, we describe the relationship between $\bm a$  and $\bm u$ such that $\mathcal{M}: \mathcal{A} \times \mathcal{U} \mapsto \mathcal{F}$, forms a triplet of Banach spaces $(\mathcal{U}, \mathcal{V}, \mathcal{F})$. A graph neural operator $\mathcal{G}$  approximates the operator $\mathcal{M}$ through a kernel integration performed by message passing on graph networks \cite{li2020multipole}. However, in contrast to the graph neural operator proposed to learn the solutions from the input, our approach involves a graph operator, which enables the transformation of input functions to latent regular space. Here, the input to the graph operator contains distance function $\bm T$; input functions $\bm s$ and boundary conditions $\bm g$, which are collectively denoted as $\bm a=\{\bm s, \bm T, \bm g\}$. 

The graph operator contains an uplifting layer $P$, multiple layers comprising point-wise and integral layer $k_{l=0:L}$, and a down-lifting layer $Q$. $P$ is a point-wise transform that lifts the original space to high dimensional space such that $P: \bm a \mapsto \bm v_{0}$. The uplifted input undergoes a series of  kernel integral operations such that at each integration operation transforms $v_{l-1}$  to  $v_l$, where the integral operation is expressed as follows:
\begin{equation}\label{kern_int}
    v_l(x) = \int_{D_{a}} \kappa_l(x, y) v_{l-1}(y) dy,
\end{equation}
where $\kappa_l$ in the above equation denotes a learnable kernel function, $v_{l-1}$  and  $v_l$ represents input and output at the $l^{th}$ level of kernel integration operation. The output of the last integral operation is projected back to the original dimension through transformation $Q$. Thus, overall, the operation can be represented as a composition operation given as follows:
\begin{equation}\label{kernel_iteration}
    \Phi(a(x)) = Q \circ K_L \circ \ldots \circ K_1 \circ P(a(x)).
\end{equation}
In a graph neural operator, the kernel integral operation in \autoref{kernel_iteration} is achieved through a mechanism called message-passing, where the graph networks employ a standard architecture utilizing edge features \cite{gilmer2017neural}. Fundamentally, the message-passing framework is devised of a graph that connects the physical domain $D_a$. The nodes in this graph are chosen as the $K$ discretized spatial locations, and edge connectivity is determined based on the integration measure, specifically the Lebesgue measure restricted to a ball \cite{li2020neural}. Each node $x \in \mathbb{R}^d$ is connected to other nodes within the ball $B(x, r)$ and forms the neighborhood set $N(x)$. For each neighbor, $y \in N(x)$, the edge weight is determined as $e(x, y) = (x, y, a(x), a(y))$. The local structure facilitates more efficient computation through a Monte Carlo approximation of the kernel integral while remaining invariant to mesh refinement. Additionally, a Nystrom approximation \cite{nystrom1930praktische} of the kernel is employed to further enhance the computation efficiency.

Having discussed the theoretical aspects of kernel integration in the graph neural operator, we now move on to the implementation. According to \cite{li2020multipole}, the integral in \autoref{kern_int} can be truncated  within a given local ball $B_r(x)$ at $x$ with radius $r > 0$,
\begin{equation}\label{kern_int_ball}
    v_l(x) = \int_{B_r(x)} \kappa(x, y) v_{l-1}(y) \, dy. 
\end{equation}
Further, sampling on the discredited space is carried out such that for each point $x \in D$, there is a set of randomly sampled points $\{ y_1, \ldots, y_M \} \subset B_r(x)$, and the sampled input mesh points are connected with a graph. In a discretized space, the integral is approximated using a Riemannian sum. Thus the integral given in the \autoref{kern_int_ball} can be approximated as:
\begin{equation}
    v_l(x) \approx \sum_{i=1}^M \kappa(x, y_i) v_{l-1}(y_i) \mu(y_i),
\end{equation}
where $\mu$ represents the Riemannian sum weights corresponding with the space defined by $ B_r(x)$. The construction of a graph operator-based encoder involves applying kernel integration to the input defined on the physical domain. For that, consider the coordinates points in the irregular physical domain  $\{x_1^{\text{in}}, \ldots, x_N^{\text{in}}\} \in D_a$, and corresponding input function points are
$\{y_1^{\text{in}}, \ldots, y_N^{\text{in}}\}$. The encoder operator transforms these input points into a corresponding function defined on uniform latent grid points $\{x_1^{\text{grid}}, \ldots, x_S^{\text{grid}}\} \in D_r$. The kernel integration over the ball $B_r(x_{\text{grid}})$  enables this encoder transformation can be expressed as:
\begin{equation}
 v_0(x^{\text{grid}}) \approx \sum_{i=1}^M \kappa(x^{\text{grid}}, y_i^{\text{in}}) \mu(y_i^{\text{in}}). 
\end{equation}
For a non-uniform grid, Riemannian sum weights $\mu(y_i^{\text{in}})$ need to be computed as they account for varying densities.
The overall operation of transforming the input on a regular grid to a latent space can be defined using a graph operator-based encoder $\hat{\mathcal{M}}_E$ such that
\begin{equation}
 v_0(\bm{x}^{\text{grid}})  = \hat{\mathcal{M}}_{E} (\bm{x}^{\text{grid}}, \bm{y}^{\text{in}}). 
\end{equation}
Similarly, we also devise a decoder operator which enables a transform of the input function on a uniform latent grid $\{x_1^{\text{grid}}, \ldots, x_S^{\text{grid}}\}$, to output function on the original space having an irregular grid. The graph operator-based decoder obtains output at coordinate points $\{x_1^{\text{out}}, \ldots, x_N^{\text{out}}\}$ The kernel integration over the ball $B_r(x_{\text{out}})$ obtains output is expressed as:
\begin{equation}
u(x_{\text{out}}) \approx \sum_{i=1}^M \kappa(x_{\text{out}}, y_i^{\text{grid}}) v_l(y_i^{\text{grid}}) \mu(y_i^{\text{grid}}). 
\end{equation}
In contrast to the Riemannian weights obtained for kernel integration in the encoder, here the weights can be obtained as $\mu(y_i^{\text{grid}}) = 1/M$ as grids of the chosen latent space are uniform, with $M$ being the distance between two grid points in the latent space. The overall operation of transforming the output on the latent space to the physical space can be defined using a graph operator-based decoder $\hat{\mathcal{M}}_D$ and the operation which can be provided as:
\begin{equation}
 u(\bm{x}^{\text{out}})  = \hat{\mathcal{M}}_{D} (\bm{x}^{\text{out}}, \bm{y}^{\text{grid}}). 
\end{equation}

\subsection{Waveformer}
Following the discussion on graph-based operators for both the encoder and decoder,  our discussion now focuses on the second component: the waveformer. We have previously discussed kernel-based neural operators, including the Graph Neural Operator (GNO), Fourier Neural Operator (FNO), and Wavelet Neural Operator (WNO). These operators share a common foundation, where the core idea is the step-wise updation of the uplifted input as shown in \autoref{kernel_iteration}. Here, the kernel of the integral transformation, parameterized by a neural network, is learned through end-to-end training. 
In contrast to the GNO, FNO, and WNO use Fourier and wavelet transforms to enable learning the kernel parameters effectively in the spectral space. Moreover, after the integral kernel operation at each step, the output is summed up with the output of a linear transformation applied to the input. Thus, overall, the step-wise update can be expressed as:
\begin{equation}\label{eq:iteration}
v_{l} =  \phi (\psi^{-1}(\left(K(\phi) \cdot \psi(v_{l-1})))(x) + W v_{l-1}(x) \right) \quad \text{for } x \in D \text{ and } l \in [1, L],
\end{equation}
where $\cdot $ denotes convolution, $\phi$ is the activation function, $K(\phi)$  represents the kernel parameterized by $\phi$, $\psi$ and $\psi^{-1}$ are the forward an d inverse spectral transforms and $W$ is a weight matrix enables the linear transformation. When a wavelet transform is used in \autoref{eq:iteration} the $\psi$ and $\psi^{-1}$ can be replaced by forward wavelet transform $\mathcal{W}(\cdot)$ and a inverse wavelet transform $\mathcal{W}^{-1}(\cdot)$, respectively,
\begin{equation}\label{eq:wavelet}
    \begin{aligned}
        (\mathcal{W} v_{j})(s, \tau) & = \int_{D} v_j(x) \frac{1}{|s|^{1 / 2}} \psi\left(\frac{x-\tau}{s}\right) dx,  \\
        (\mathcal{W}^{-1} (v_{j})_w)(x) & = \frac{1}{C_{\psi}} \int_{0}^{\infty} \int_{D} (v_{j})_{w}(s, \tau) \frac{1}{|s|^{1 / 2}} \tilde{\psi}\left(\frac{x-\tau}{s}\right) d\tau \frac{ds}{s^{2}}.
    \end{aligned}
\end{equation}
In the above equations, $\psi(x)$ represents the orthonormal mother wavelet, while $s$ and $\tau$ are the scaling and translational parameters involved in wavelet decomposition. The term $(v_{j})_{w}$ denotes the wavelet decomposed coefficients of $v_j(x)$. $\tilde{\psi}(\cdot)$ represents the scaled and shifted mother wavelet, and $0 < C_{\psi} < \infty$ is an admissible constant as defined by Daubechies \cite{daubechies1992ten}.

Waveformer differs from the aforementioned operator learning frameworks, where kernel integration is performed through element-wise multiplication of the learnable parameters. Waveformer redefines the kernel operation and step-wise updation by performing nonlinear integration employing a transformer. The operation is expressed as:
\begin{equation}\label{eq:trans1}
    v_{out1}(x) = \mathcal{W}^{-1}(T_{\mathcal{W}}(\mathcal{W}(v_{in}(x)))).
\end{equation}
where $v_{in}$ denotes the uplifted input, $T_{\mathcal{W}}$ represents the transformer acts on the wavelet domain and $v_{out1}$ is the corresponding output. Similarly, the action of transformer $T_{R}$ obtains a nonlinear transformation on the physical space, effectively replacing the series of activation and linear transformations described  in \autoref{eq:iteration},
\begin{equation} \label{eq:trans2}
    v_{out2}(x) = T_{\mathcal{R}}(v_{in}(x)).
\end{equation}    
Here $v_{out2}(x)$ is the output of the transformer $T_{R}$. 
Hence, the final output yields from the waveformer can be expressed as the summation of the outputs $v_{out1}$ and $v_{out2}$ given below:
\begin{equation}\label{eq:summ}
    v_{out} = G_{T}(v_{in}) = {\mathcal{W}^{-1}}{(T_{\mathcal{W}})}{({v_{in}}(x))}+T_{R}({v_{in}}(x))
\end{equation}
It is emphasized that while diverse families of wavelets exist with distinct characteristics, we specifically employ the Daubechies wavelets in the current work. The Daubechies wavelets are known for their simplicity, compactness, orthogonality, and resilience in texture retrieval tasks and adaptability to different levels of smoothness \cite{daubechies1992ten}. The employed transformer consists of a series of stacked encoder and decoder blocks. The encoder employs self-attention, while the decoder utilizes both self-attention and encoder-decoder cross-attention. In encoder-decoder cross-attention, the attention scores are computed between the query of the decoder input and the keys of the encoder embeddings. The self-attention mechanism ensures a seamless kernel integration within the wavelet domain, whereas the encoder-decoder attention maintains the continuity of temporal information. The figure provides the schematic description of the architecture of the waveformer.
\begin{figure}[t!]
    \centering
    \includegraphics[width=\textwidth]{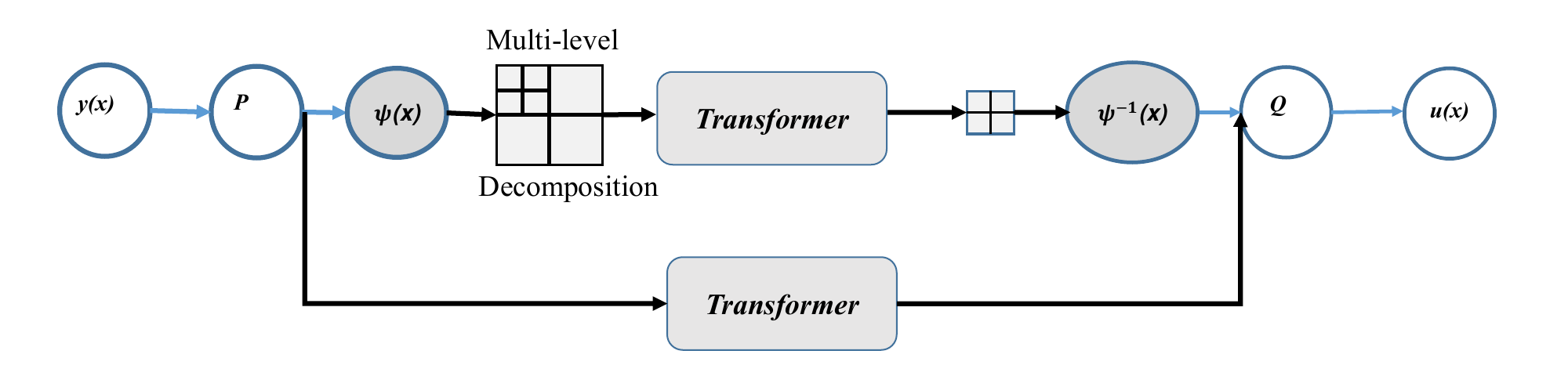}
    \caption{\textbf{A diagrammatic representation of the waveformer architecture}. Firstly, inputs are uplifted by passing through local transformation $P$. The integral layer consists of two separate branches. In the first branch, wavelet decomposition is performed on the uplifted images and then is passed through a transformer. In the second branch, the inputs are directly fed to the other transformer. An activation is applied in the resultant output obtained by summing the outputs of the two branches. Then, the outputs are downlifted by passing through the transformation $Q$, which yields the prediction u(x). Here, the local transformations $P$ and $Q$ are modeled as fully connected neural networks of one hidden layer.}
    \label{fig:architectur2}
\end{figure}

\subsection{Geometry adaptive waveformer}\label{subsec:geom_adap_wf}
\begin{figure}[ht!]
    \centering
    \includegraphics[width=\textwidth]{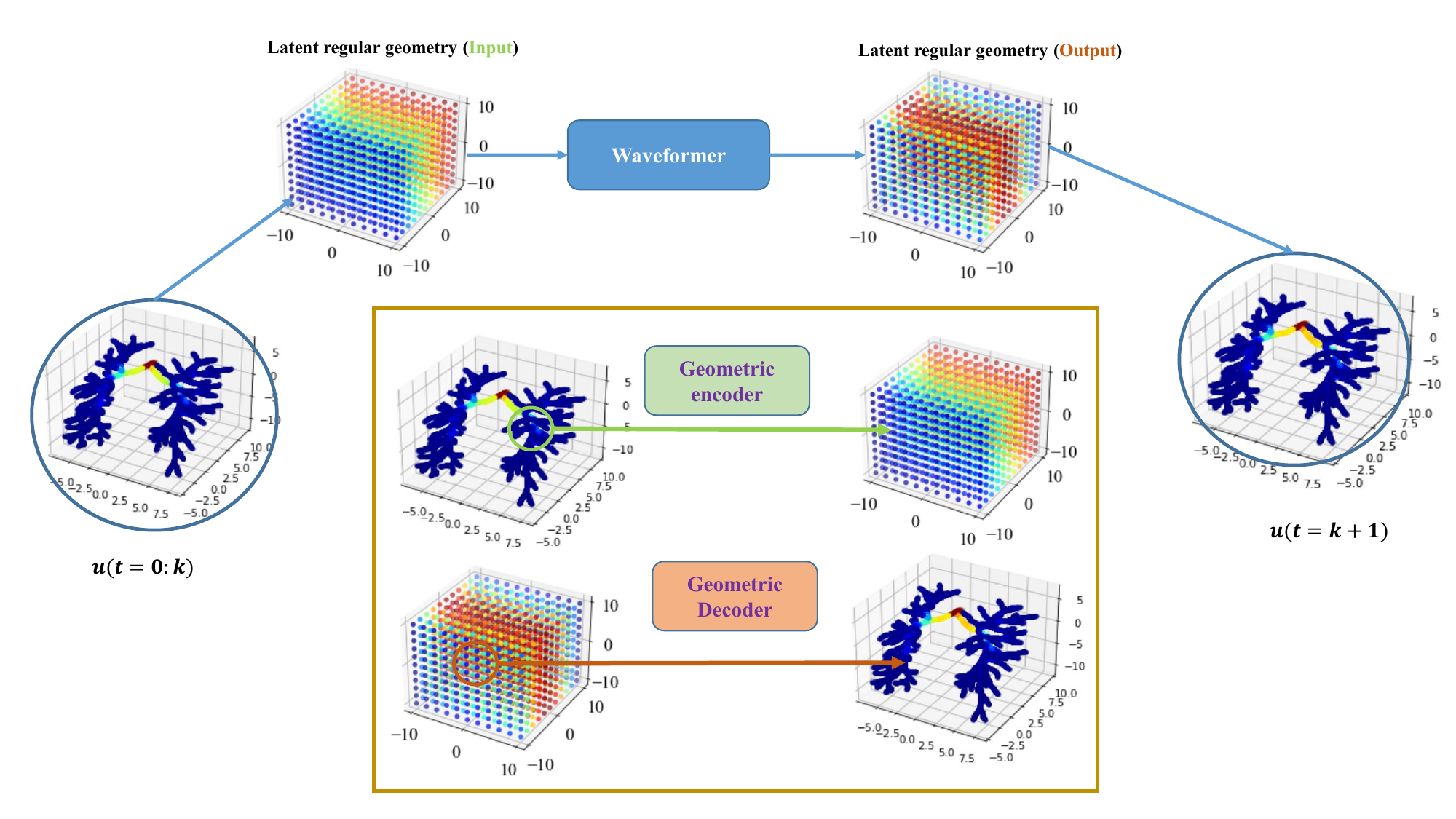}
    \caption{Schematic of the geometry-adaptive waveformer, which consists of three components: a geometry encoder, a waveformer, and a geometry decoder. The initial input, $a_0(x)$, is passed through the geometry encoder, which transforms the input into a regular latent grid. The waveformer then operates on this latent space, and finally, the geometry decoder maps the output back to the original space to predict the response at the subsequent time step $u_{k+1}$.}
    \label{fig:gwaveformer}
\end{figure}

The geometry adaptive waveformer is formulated by integrating three primary components: a geometry encoder, a waveformer, and a geometry decoder. A schematic of the proposed framework is shown in \autoref{fig:gwaveformer}. We employ the framework to analyze and simulate cardiovascular systems. To investigate the set of $N$ patient-specific cardiovascular geometries, denoted as $g_1, g_2, \ldots, g_N$, we use $N$ distinct geometry adaptive waveformer models. Each model obtains the prediction of a discrete-time response sequence $u_0, u_1, \ldots, u_n$ at the time steps  $t_0, t_1, \ldots, t_n$, where the constant time increment is defined as $\Delta t = t_1 - t_0 = \ldots = t_n - t_{n-1}$. These time responses represent the evolution of parameter fields (such as pressure and velocity) across the specified geometry during a complete cardiac cycle. While the original waveformer operates within a 3D latent space, it can lead to an increase in model parameters due to the high dimensionality. A modified version has been developed to address the challenge and to enhance the efficiency of the originally proposed geometry-adaptive waveformer.
The modified geometry-adaptive waveformer achieves efficiency by introducing two convolutional neural networks (CNN)--based blocks: a reduction block and an expansion block. The reduction block gradually transforms the 3D latent space into a 2D latent space, effectively reducing the dimensionality. This reduction is achieved through a Reduction block consisting of sequence of 3D convolutional operations, as described by: $\text{Conv3D}(\text{latent 3D}, \text{kernel-1R}, \text{channel-1R}) \mapsto \text{Conv3D}(\text{Dim R}, \text{kernel-2R}, \text{channel-2R}) \mapsto \text{Flatten} \mapsto \text{Reshape}(\text{latent 2D}, \text{channel-3R})$.

\begin{figure}[ht!]
    \centering
    \includegraphics[width=\textwidth]{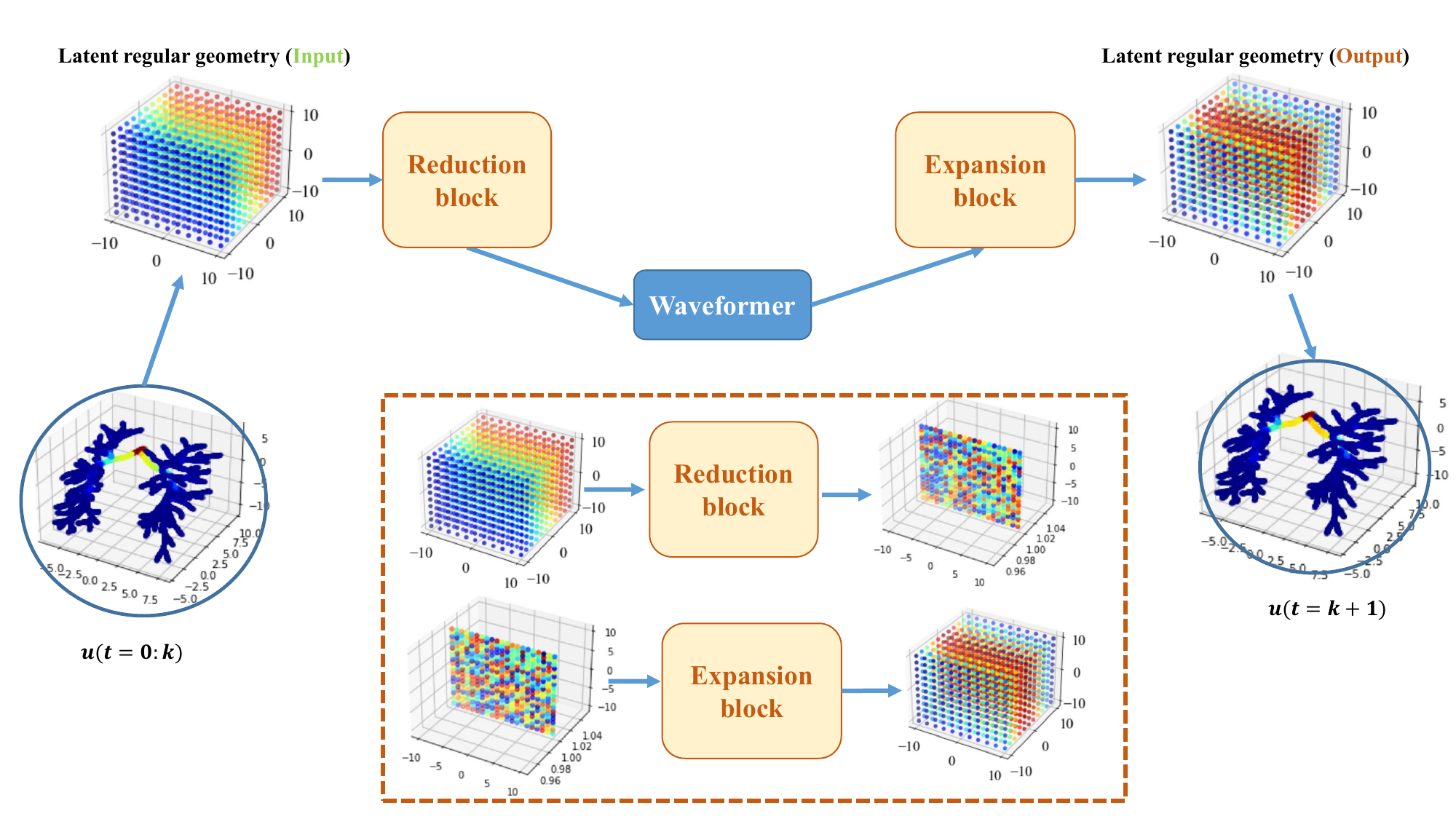}
    \caption{Schematic of modified geometry-adaptive waveformer, which consists of three components: a geometry encoder, a waveformer, and a geometry decoder. The initial input, $a_0(x)$, is passed through the geometry encoder, which transforms the input into a regular latent grid. The reduction block reduces the dimension of the 3D latent space to 2D space. The waveformer then operates on this 2D latent space and, subsequently, transforms it into a 3D latent space employing an expansion block. Finally, the geometry decoder maps the output back to the original space to predict the response at the subsequent time step $u_{k+1}$.}
    \label{fig:modifiedgwaveformer}
\end{figure}

Once the latent space is reduced to 2D, the 2D waveformer is employed to learn the dynamics efficiently. Following this, the expansion block maps the reduced 2D latent space back to the original 3D space. This transformation is achieved through a series of transposed 3D convolutional operations. The operation of this Expansion block is as follows: $\text{Flatten} \to \text{Reshape}(\text{latent 3D}, \text{channel-1E}) \to \text{ConvTranspose3D}(\text{Dim E}$ $, \text{kernel-2E}, \text{channel-2E}) \to \text{ConvTranspose3D}(\text{latent 3D}, \text{kernel-3E}, \text{channel-3})$ .The schematic of the overall architecture of the modified geometr adaptive waveformer is shown in \autoref{fig:modifiedgwaveformer}.

The proposed geometry adaptive waveformer is devised as an operator learning paradigm that aims to predict the future spatiotemporal responses from the initial spatiotemporal response observed over limited time steps for any generic geometry. Let us consider $\bm{u}_{k:k+n}$ as the output response from time-step $k$ to a time-step $k+n$, and initial responses till $k$ time steps $\bm{u}_{0:k}$ with $d_0$ state variables over the discretized domain $D_a$. The discretization on a domain $D_a$ is done with $D_i$ points in the $i$-th dimension such that for 3D, $\bm{u}^n \in \mathbb{R}^{d_0 \times D_1 \times D_2 \times D_3}$. The initial input to the geometry adaptive waveformer is given by $\bm a_0(\bm x) = \bm u_{0:k}$, and it obtains the prediction of response at the very next time step $u_{k+1}$. Within the framework, the input $\bm a_0(\bm x)$ is first mapped on to $\bm x^{grid}$, which yields $\bm a^{g}_0(\bm x)$ such that $\bm a^{g}_0(\bm x) = \hat{\mathcal{M}}_{E}(\{{\bm a}_0(\bm x),\bm x,T(\bm x)\})$, where $\bm T(\bm x)$ is signed distance function features. Here, it is noteworthy that the transformer is provided with two inputs, one for the encoder and the other for the decoder, to enable the learning sequence. Thus we split the initial input, $\bm a^g_{0}(x)$ into two parts $\bm a^{g}_{0_{enc}}= \bm u^{g}_{0:k-1}$ and $\bm a^{g}_{0_{dec}} = \bm u^{g}_{1:k}$. Passing the inputs to the waveformer gives the output $u^{g}_{k}$ such that $u^{g}_{k+1} = \hat{\mathcal{M}}_{T}(\bm u^{g}_{0:k-1},\bm u^{g}_{1:k})$. Further, it is transformed into the physical domain $\bm u_{k+1}(\bm x) = \hat{\mathcal{M}}_{D}(\{\bm u^{g}_{k},\bm x^{grid}, \bm T(\bm x^{grid})\})$. Following the prediction of the response at $({k+1})^{th}$ time step, input in the next time step can be updated as $\bm a_{0}(x)=\{\bm a_{0_{1:k}}(x),\bm u_{k+1}\}$. Similarly, input for the subsequent time step is updated as $\bm a_{0}(\bm x)=\{\bm a_{0_{2:k+1}}(\bm x),\bm u_{k+2}\}$. Thus, the general updation scheme at $n^{th}$ time step is expressed as:
\begin{equation}
    \bm a_{0}(\bm x)=\{\bm a_{0_{n:n+k-1}}(\bm x),\bm u_{n+k}\}.
\end{equation}
The output prediction $\bm u_{n}$ at the $n^{th}$ time step can be obtained by employing the aforementioned roll-out scheme recursively. A training algorithm for geometry waveformer is given in algorithm \ref{algo:GAdwaveformer} where the forward passing through the model is given in algorithm \ref{algo:prediction}.
\begin{algorithm}[ht!]
    \caption{Training algorithm for geometry adaptive waveformer}\label{algo:GAdwaveformer}
    \begin{algorithmic}[1]
    \Require{$N$-number of paired input and output $\left\{\bm a(\bm x) = \{\bm s,\bm T, \bm g\} \in \mathbb{R}^{n_D \times d_a}, \bm u(\bm x) \in \mathbb{R}^{n_D \times d_u}\right\}$ and coordinates $\bm x \in D$}
         \State {\textbf{Initialize:} Set of trainable network parameters, $\bm{\theta} = \{\bm{\theta}_{E},\bm{\theta}_{D},\bm{\theta}_{T}\} $} of the geometry encoder, decoder and waveformer.
         
        \State{Stack the inputs: $\{\bm a(x), \bm x\} \in \mathbb{R}^{n_D \times (da+1)}$.}
        
        \For{for epoch $=1, \ldots$, epochs}
            \State{Perform the geometry encoding transformation on input $\mathcal{M}_{E}(\cdots,\bm{\theta}_{E}): \{a(x), x\} \mapsto \bm{y}_{in}(\bm{x}_{grid})$}
            
            \State{Employ the waveformer $\mathcal{M}_{T}(\cdots,\bm{\theta}_{T}): \{\bm{y}_{in}(\bm{x}_{grid}), x^{grid}\} \mapsto \bm{y}_{out}(\bm{x}_{grid})$ }
            
            \State {Perform the geometry decoding transformation on input $\mathcal{M}_{D}(\cdots,\bm{\theta}_{D}): \{\bm{y}_{out}(\bm{x}_{grid}), \bm x^{grid}\} \mapsto \hat{\bm{u}}(\bm{x})$}

            \State{Continue the forward passing to obtain the response at the future time steps} 
            
            \State{Compute the reconstruction loss:$\mathcal{L}(\bm \theta) = \|\bm{u}(x),\hat{\bm u}(x)\|^2_p$}
            
            \State{Update the trainable parameters : 
             $\bm{\theta} \leftarrow \bm{\theta} -\delta \nabla_{\bm \theta} \mathcal{L}(\bm \theta)$}
        \EndFor
    \Ensure{Output prediction of waveformer $\bm u \in \mathcal{U}$ for the optimal network parameters}
    \end{algorithmic}
\end{algorithm}

\begin{algorithm}[ht!]
    \caption{Prediction of geometry waveformer through forward passing}\label{algo:prediction}
    \begin{algorithmic}[1]
    \Require{Geometry waveformer model: $\mathcal{M}$; Input:$\bm a_{0}(\bm x)$; Number of time-steps to predict in future:$ m$}
        \State {\textbf{Initial inputs to the model:} $\bm a_{0}(\bm x)=\bm u_{0:k}$ and grid $\bm x$.}
        \For{for i $=1, \ldots$, $\bm m$}
            \State{Stack the input and the grid: $\bm u_{in} \leftarrow \{\bm a_{0}(\bm x), \bm x\}$.}
            \State{Predict the output: $\bm u_{k+i} \leftarrow \mathcal{M}(\bm{u}_{in};{\bm \theta})$}
            \State{Update the input : $\bm a_{0}(\bm x)=\{\bm a^{0}_{1:k}(\bm x),\bm u_{k+i}\}$}
        \EndFor
    \Ensure{Prediction of response from time steps $k$ to $m$, $\bm u_{k:m}$}
    \end{algorithmic}
\end{algorithm}

\begin{remark} \label{training scheme 2}
Algorithm (\ref{algo:prediction}) for the geometr adaptive waveformer predicts subsequent time steps using responses from $k$ initial time steps. While this algorithm effectively predicts future time steps, the roll-out begins with the initial responses for the $k$ time steps. Considering the realistic scenario, where only the initial states are known prior, a modified version of the scheme is introduced here to enable the model to progressively predict directly from the initial conditions, $\bm{u}_0$, to the end time of the cycle. For the given geometry-adaptive waveformer, $\mathcal{M}$, the system's response over $n$ time steps is predicted in an auto-regressive manner. The progressive prediction from the initial conditions, $\bm{u}_0$, to subsequent time steps is performed as follows:
\begin{equation}
\bm u_1 = \mathcal{M}(a_{0_1}; \bm \theta), \quad \text{where} \ \bm a_{0_1} = \{\bm u_0, \bm u_0, \ldots, \bm u_0, \bm u_0\}
\end{equation}
\begin{equation}
\bm u_2 = \mathcal{M}(a_{0_2}; \bm \theta), \quad \text{where} \ \bm a_{0_2} = \{\bm u_0, \bm u_0, \ldots, \bm u_0, \bm u_1\}
\end{equation}
\begin{equation}
\vdots
\end{equation}
\begin{equation}
\bm u_n = \mathcal{M}(\bm a_{0_n}; \bm \theta), \quad \text{where} \ \bm a_{0_n} = \{\bm u_{n-1-k},\bm u_{n-1-k}, \ldots,\bm u_{n-2}, \bm u_{n-1}\}
\end{equation}
$k$ length of the input $\bm a_{0}$. To predict the response at $n^{th}$ time $\bm u_{n}$, the model is run $n$ times string from zero.
\end{remark}

\section{Results and discussion}\label{sec: Numerical example}
In this section, we illustrate the performance of the proposed approach in solving cardiovascular flow. We consider four different scenarios involving the healthy pulmonary model, Aorta model affected by coarctation, and the healthy aorta models, and the objective is to learn the pressure and the flow fields using the proposed geometry adaptive waveformer. To evaluate the performance of the proposed approach, we report the relative mean squared error and present a scatter plot. We also present contour presents comparing the pressure and flow-fields obtained using the proposed approach ground truth data generated using numerical simulation. Lastly, the validated model is employed to quantify the propagation of uncertainty from the initial condition to the pressure and the flow fields. 

The proposed architecture consists of the encoder operator, decoder operator, and the waveformer. The encoder and decoder operators each consist of two graph convolution layers with hidden dimensions of 64 and 32, respectively. The dimensions of the input and output nodes are devised such that they match the respective input and output feature dimensions of the dataset. The waveformer consists of a single wavelet decomposition layer, which is followed by a transformer consisting of 2 encoder blocks and 2 decoder blocks. In order to optimize the network parameters, the ADAM optimizer with an initial learning rate varying from $10^{-4}$ to $10^{-3}$ is used. A learning rate decay of 0.6 is scheduled to enhance the convergence during the training, where the learning rate is reduced by a factor of 0.6 every 5 epochs. The model is trained for 100 epochs with a batch size of 1.

\subsection{Dataset}
The dataset utilized for the study comprises 4 cardiovascular models from the Vascular Model Repository (VMR) {\color{red}[ref]}, including 3 healthy and 1 affected aorta. These models were purposefully chosen for their complex features, such as multiple junctions and stenoses, which, in general, pose a challenge to traditional models. The data-generation pipeline involves two major steps: firstly, three-dimensional simulations were performed with varying boundary conditions over two cardiac cycles using the unsteady Navier-Stokes equations. Each simulation applied random perturbations to the boundary conditions to create diverse datasets. Post-processing focused on the model centerlines, integrating pressure and velocity over cross-sections to calculate average pressure, flow rate, and cross-sectional area. The dataset is available in the \href{https://drive.google.com/drive/folders/1UyKOAFPu8JFYH28rQ33CWI7D7CZp1aZ9}{repository}, and comprises of a total 8 simulated data. These include simulated data of five healthy aorta, a healthy pulmonary, an aortofemoral affected by an aneurysm, and an aorta model affected by coarctation. For our analysis, we utilized datasets of two healthy aorta models, one healthy pulmonary vascular model, and one aorta model affected by coarctation.

We employ distinct geometry-adaptive waveformer models tailored to each dataset, with separate models dedicated to predicting pressure and flow rate. Training error and the testing errors post-convergence are reported in the \autoref{Test-train error for the pressure} and \autoref{Test-train error for the flowrate}. A relative Mean Square Error (MSE) is used here as the metric of error evaluation. Further, we carry out a detailed evaluation of the performance of the proposed framework individually for each data set.
\begin{table}[ht!]
    \centering
    \caption{Performance of the geometric adaptive waveformer over datasets of pressure fields corresponding to different artery models.}
    \label{Test-train error for the pressure}
\begin{tabular}{lcccc} 
\hline
\textbf{Data set} & \textbf{Train error}  & \textbf{Test error}\\\hline
Healthy pulmonary model &   0.879\%	& 0.863\% \\
Aorta model affected by coarctation &	3.330\% & 1.640\%\\
Healthy aorta model 1 &	0.800\% &	0.505\%\\
Healthy aorta model 2 &	1.692\% &	1.080\% \\
\hline 
\end{tabular}
\end{table}
\begin{table}[ht!]
    \centering
    \caption{Performance of the geometric adaptive waveformer over datasets of pressure fields corresponding to different artery models.}
    \label{Test-train error for the flowrate}
\begin{tabular}{lcccc} 
\hline
\textbf{Data set} & \textbf{Train error}  & \textbf{Test error}\\\hline
Healthy pulmonary model &  0.168\% &  0.156\%\\
Aorta model affected by coarctation &  0.463\% & 0.241\% \\
Healthy aorta model 1 &  0.189\% &	0.168\%\\
Healthy aorta model 2 &  0.507\% &	0.349\% \\
\hline 
\end{tabular}
\end{table}

\subsection{Healthy pulmonary model} 
As the first dataset, we consider the healthy pulmonary model. The goal here is to learn geometry adaptive waveformer $\mathcal{M}$ that maps the spatiotemporal cardiovascular response over an initial sequence of $k$  time steps, $u_{0:k}(x) $, to the spatiotemporal responses over the subsequent $n$ time steps, $u_{k:k+n}(x)$, where the responses are defined at 18890 spacial locations, which are non-uniformly discretized across the $x$,$y$, and $z$ coordinates. The model is trained with 45 trajectories of the vascular responses and evaluated on 5 unseen trajectories. For training, initial time steps and future steps are considered to be $k=10$ and $n=20$.  The model is trained across two-time regimes: the first from time steps 10 to 30 and the second from 40 to 60. It is important to note that the responses for the intervals 30 to 40 and 60 to 80 are obtained through extrapolated predictions. 

\begin{figure}[!ht]
    \centering{
    \includegraphics[width=1.0\textwidth]{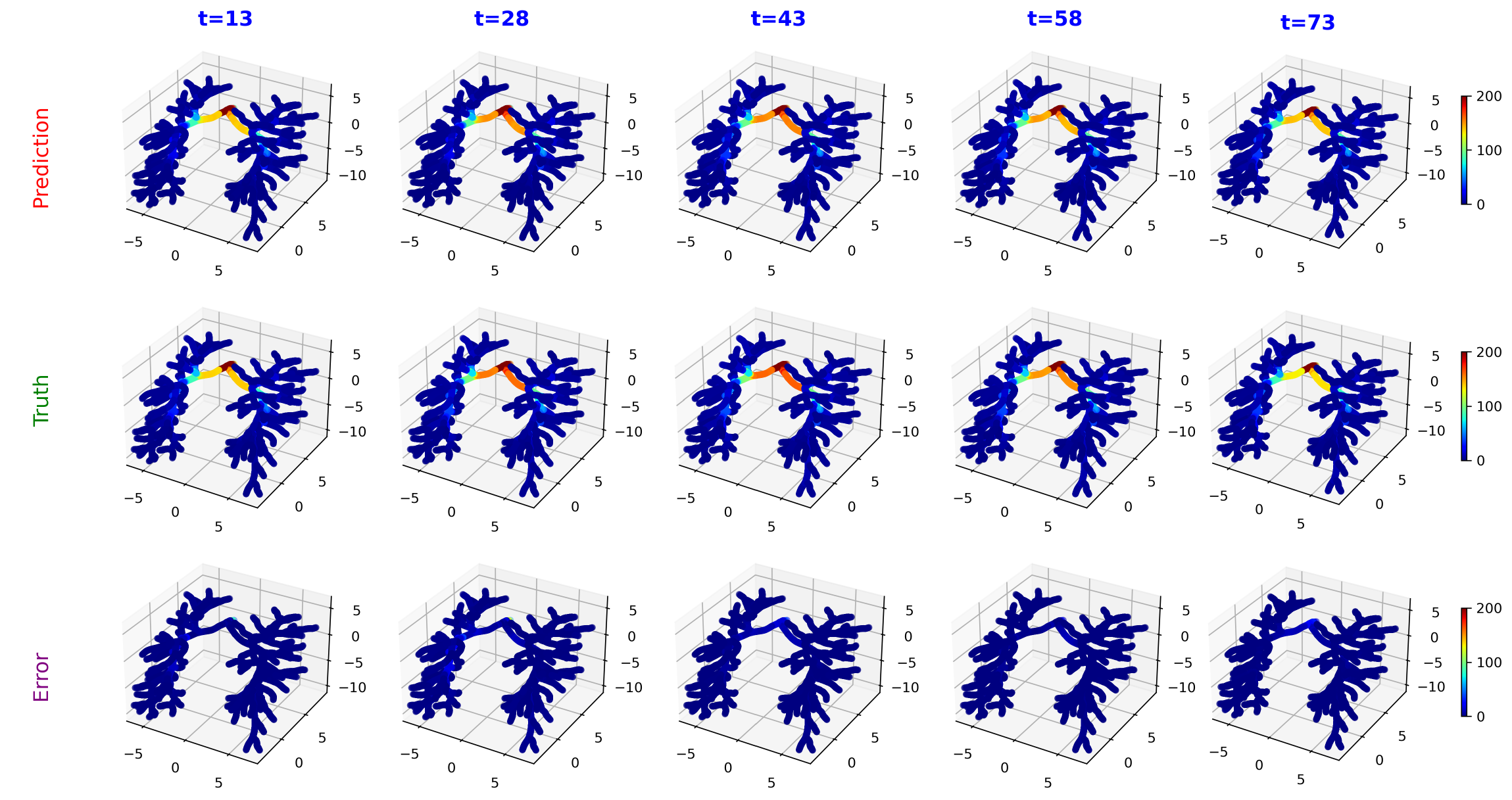}}
    \caption{Predictions results of pressure field measured in $mmHg$ for the dataset 1: (Top to bottom) Geometry adaptive waveformer prediction, ground truth response, and $L_1$ error.}\label{case1:pressure}
\end{figure}
\begin{figure}[!ht]
    \centering{
    \includegraphics[width=1.0\textwidth]{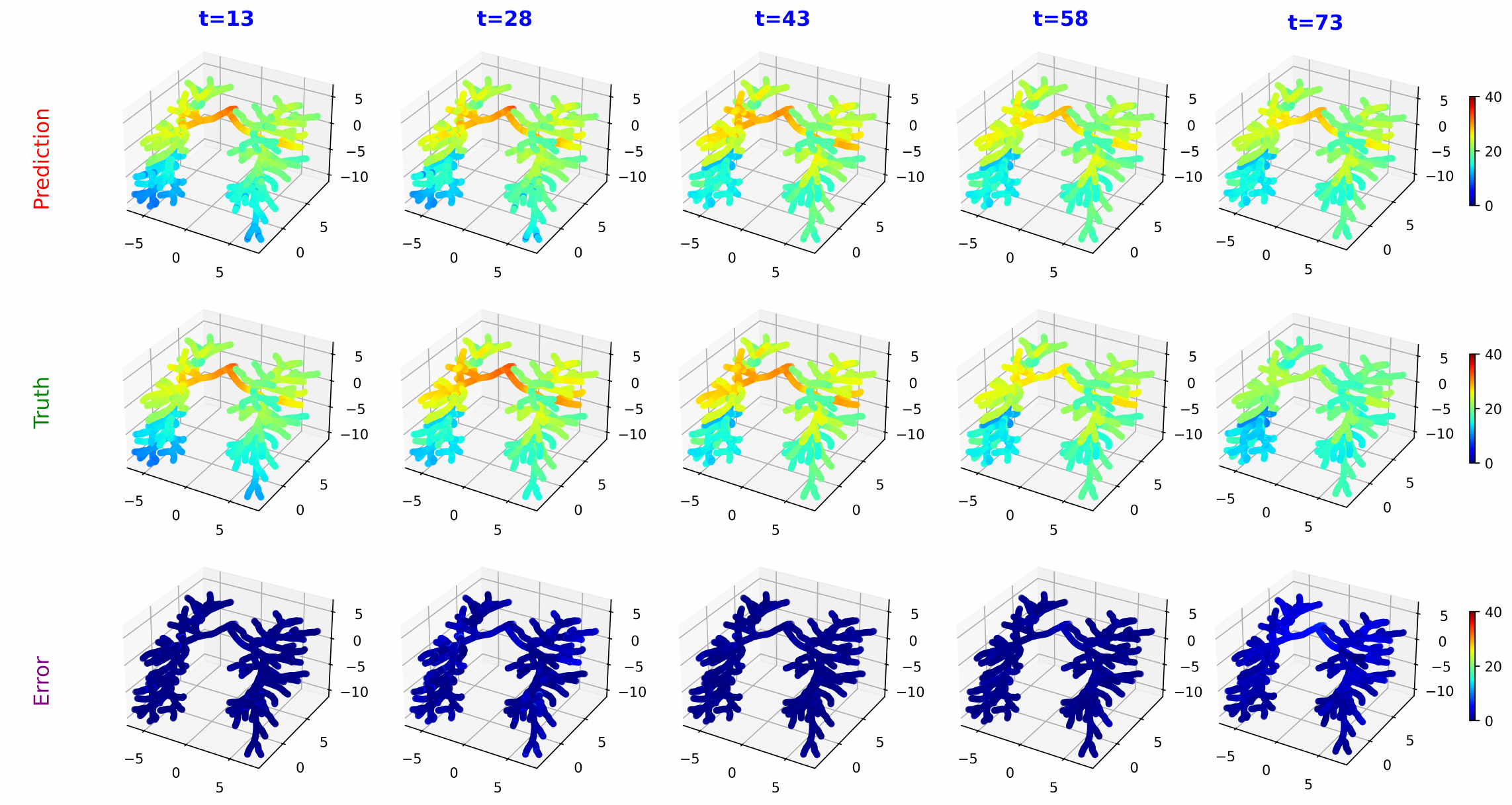}}
    \caption{Predictions results of flow rates measured in $cm^{3}/s$ for the dataset 1: (Top to bottom) Geometry adaptive waveformer prediction, ground truth response, and $L_1$ error.}\label{case1:flowrate}
\end{figure}
\begin{figure}[!ht]
    \centering
    \subfigure[]{
    \includegraphics[width=.35\textwidth]{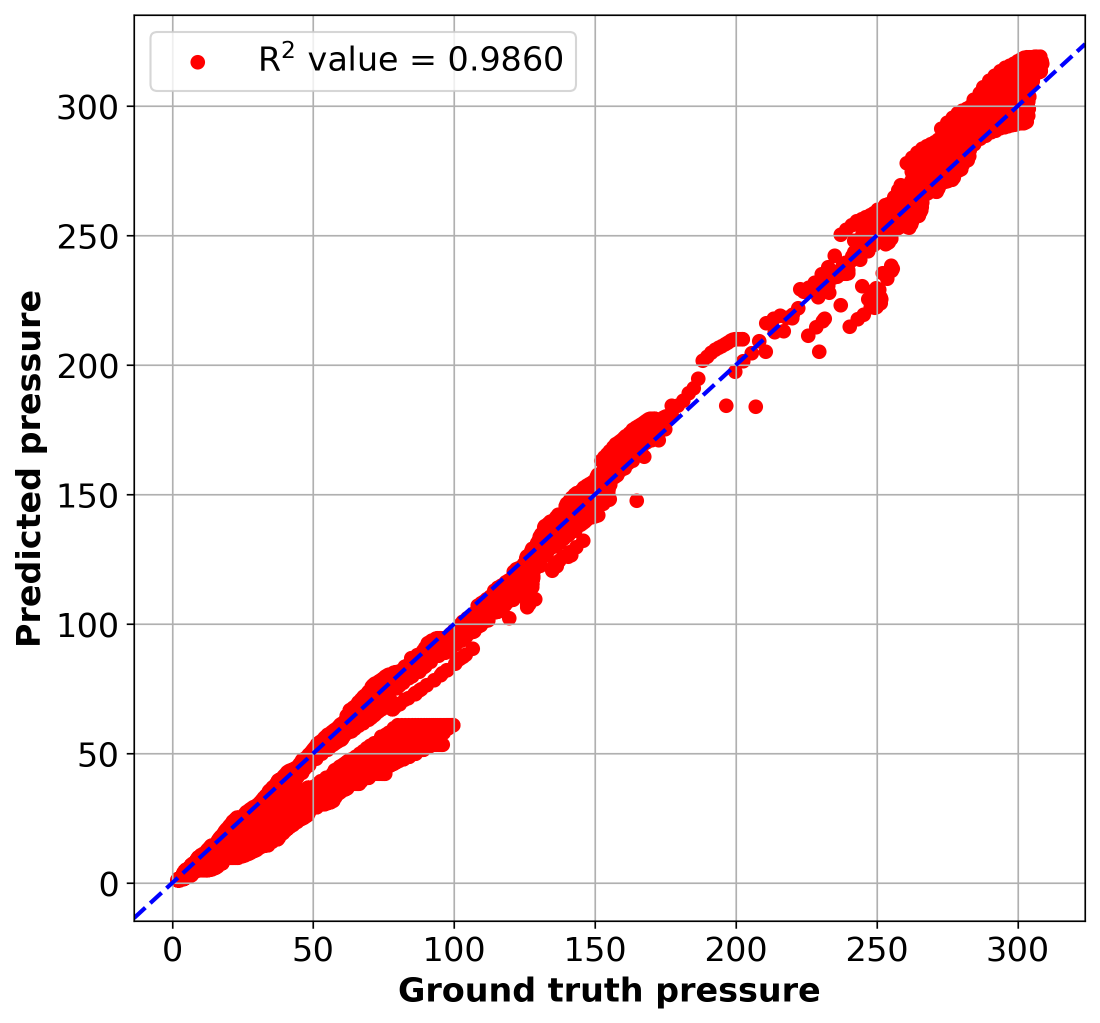}}
    \subfigure[]{
    \includegraphics[width=.35\textwidth]{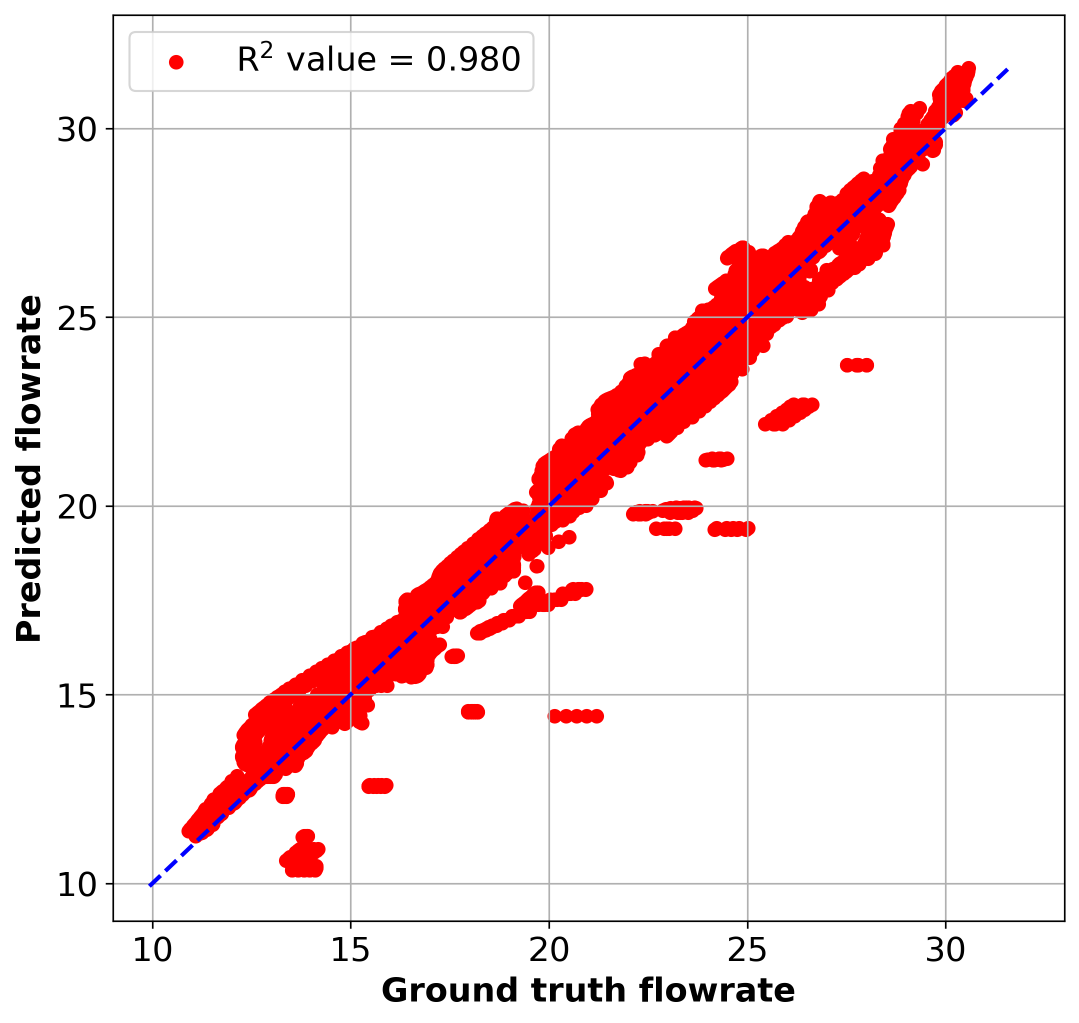}}
    \caption{Scatter plots showing the correlation between the predicted response and the ground truth for dataset 1:(a) Pressure, (b) Flow rate}
    \label{scatter1}
\end{figure}

The prediction results are presented in Figs. \ref{case1:pressure} -- \ref{scatter1}. While \autoref{case1:pressure} showcases 3-D visualization of the predicted response of pressure fields at distinctive time steps with ground truth response, \autoref{case1:flowrate} displays corresponding results of the flow rates. It is clearly visible from the results that the 3-D rendering of the predictions closely matches the ground truth. The scatter plot illustrated in \autoref{scatter1} demonstrates a strong correlation between the prediction and truth with R-squared values $0.9860$ and $0.980$ for pressure and flow rates, respectively.
\begin{figure}[!ht]
    \centering{
    \includegraphics[width=0.6\textwidth]{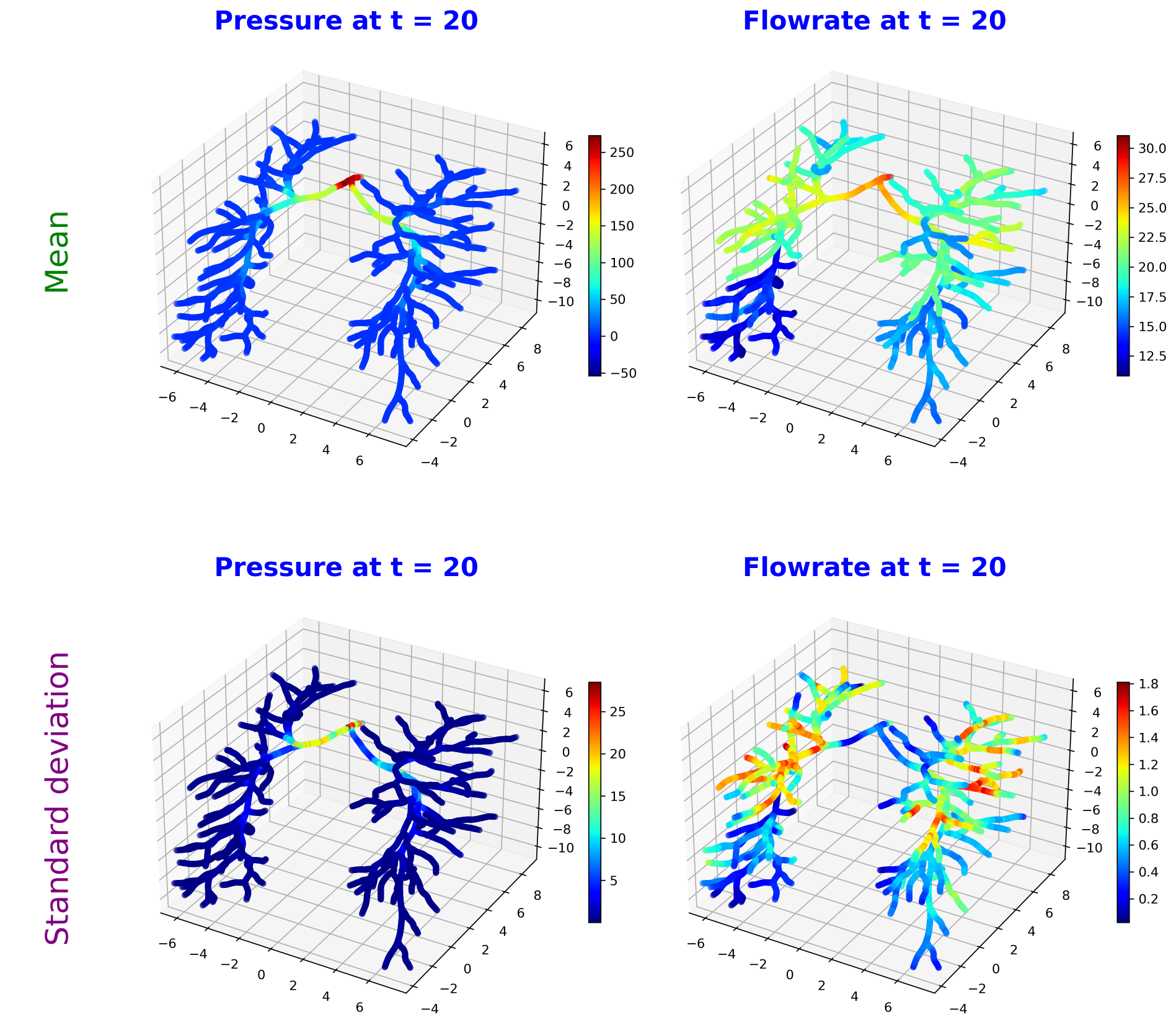}}
    \caption{Mean and standard deviation of the predicted responses of the healthy pulmonary model under uncertainty in the initial conditions.}\label{case1:mean_std1}
\end{figure}
\begin{figure}[!ht]
    \centering{
    \includegraphics[width=0.6\textwidth]{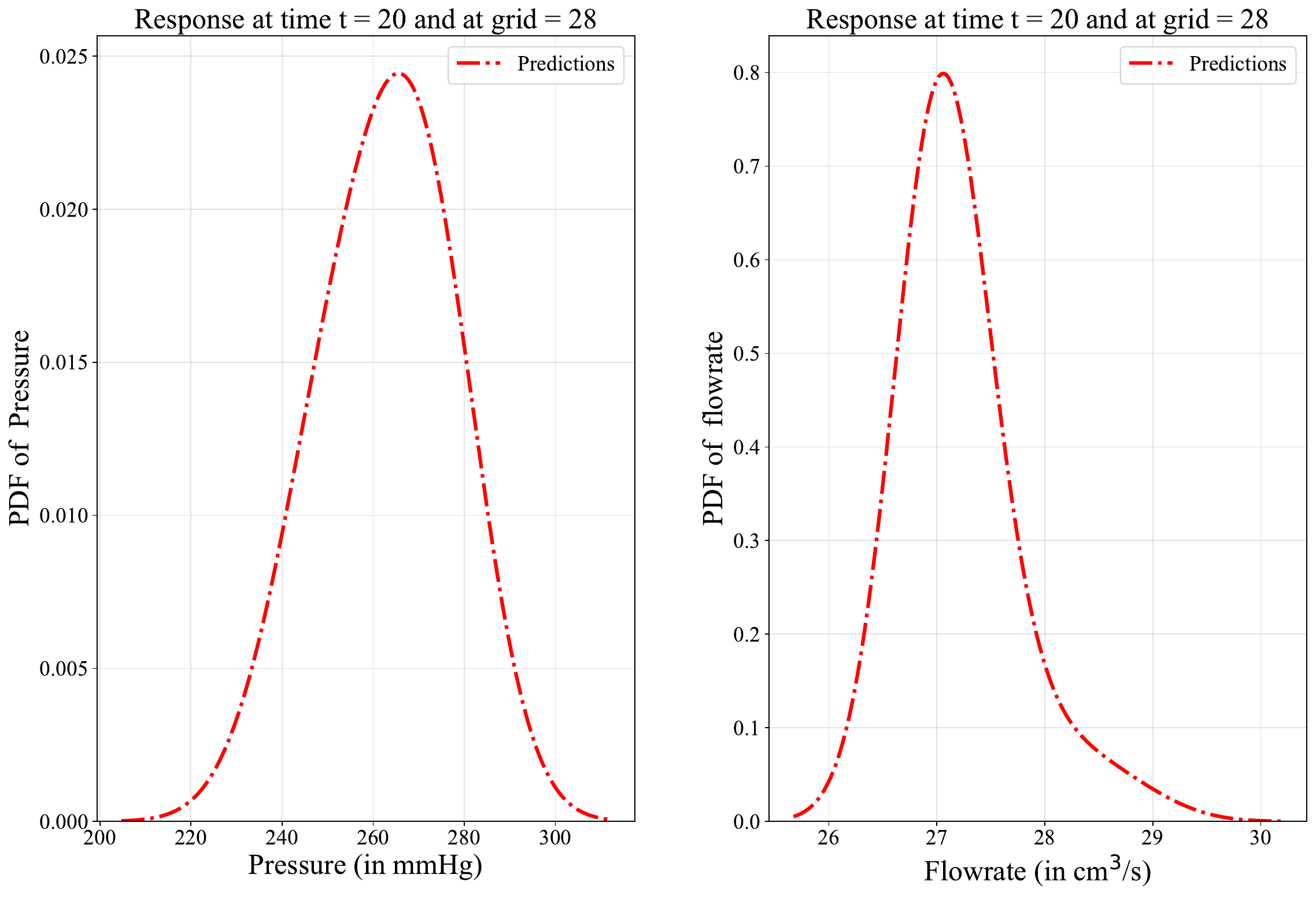}}
    \caption{Response PDF of the pressure and flowrate fields of healthy pulmonary model}\label{case1:pdf}
\end{figure}
%
%
Having established the accuracy of the geometry-adaptive waveformer, the same is employed to estimate uncertainty in the pressure and flow fields. In general, the initial and boundary conditions in cardiovascular modeling are often not known apriori. To that end, one potential solution is to model the initial and/or boundary conditions as uncertainty and quantify its effect on the response variables. In this example, the initial condition is considered to be uncertain and perturbed by adding Gaussian noise equivalent to 1\% of the standard deviation of the initial conditions. 
The model is trained to predict the solution using a time-marched scheme, starting from the initial response and progressing through subsequent time steps till the responses at the given time step. The results are presented in \autoref{case1:mean_std1} and \autoref{case1:pdf}, where \autoref{case1:mean_std1} illustrates the mean and standard deviation of the pressure and velocity fields at a specific time instant, and \autoref{case1:pdf} depicts the probability density plots of the responses at a given time and spatial location. 
{It can be observed from the results that, in the healthy pulmonary artery model, the high-pressure regions are located near the main artery branches and bifurcations. These areas experience elevated pressure due to increased flow resistance caused by the branching geometry, where blood flow splits into smaller vessels, leading to localized pressure buildup. In the mean flow rate plot, the high flow rate regions are located mostly in the central arterial pathways, which correspond to the main arteries. The higher flow rates experienced in the locations are due to their larger diameters and lower overall flow resistance compared to other smaller branches. The standard deviation plots in \autoref{case1:mean_std1} elucidate variability in pressure and flow rate. From the plots, higher variability of the pressure near the bifurcations and along certain connecting branches is observed; this is likely due to the dynamic redistribution of flow as it splits into smaller branches. In the case of the flow rate, it can be interpreted from the standard deviation that high variability is more widespread and predominantly occurs at branching points and smaller vessels. Smaller vessels tend to amplify flow fluctuations due to their higher sensitivity to disturbances and greater influence of wall resistance.}

\subsection{Aorta model affected by coarctation}
In the second case, we consider the aorta model affected by coarctation. Similar to the previous case, we aim to map spatiotemporal cardiovascular responses from an initial sequence of $k=10$ time steps to the next $n=20$ time steps. The responses are defined at 5743 non-uniformly discretized 3-dimensional spatial locations. Training and testing sets contain 45 and 5 trajectories of the vascular responses. 
\begin{figure}[!ht]
    \centering{
    \includegraphics[width=1.0\textwidth]{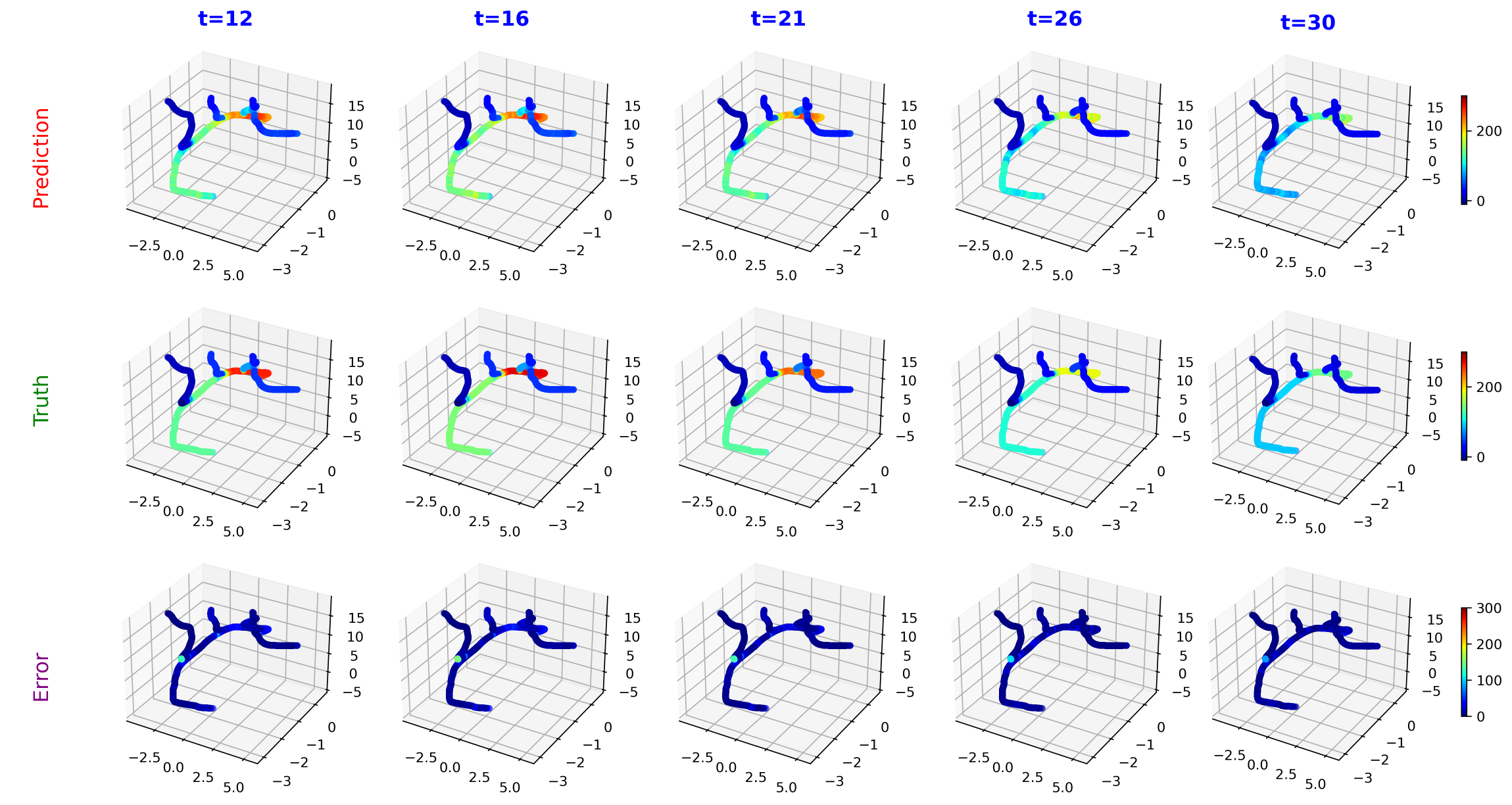}}
    \caption{Predictions results of pressure field measured in $mmHg$ for the dataset 2: (Top to bottom) Geometry adaptive waveformer prediction, ground truth response, and $L_1$ error.}\label{case2:pressure}
\end{figure}
\begin{figure}[!ht]
    \centering{
    \includegraphics[width=1.0\textwidth]{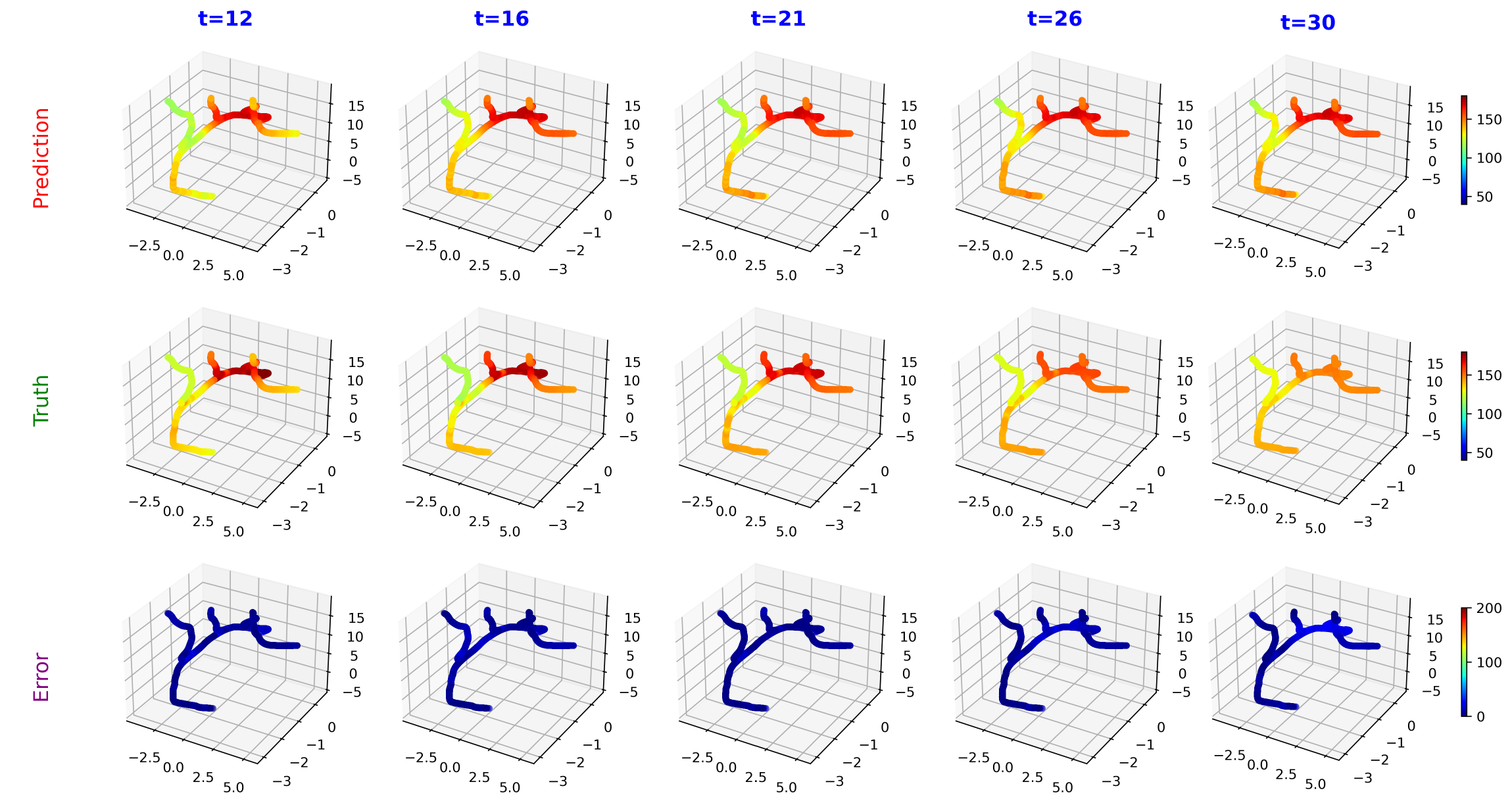}}
    \caption{Predictions results of flow rates measured in $cm^{3}/s$ for the dataset 2: (Top to bottom) Geometry adaptive waveformer prediction, ground truth response, and $L_1$ error.}\label{case2:flowrate}
\end{figure}

\begin{figure}[!ht]
    \centering
    \subfigure[]{
    \includegraphics[width=.35\textwidth]{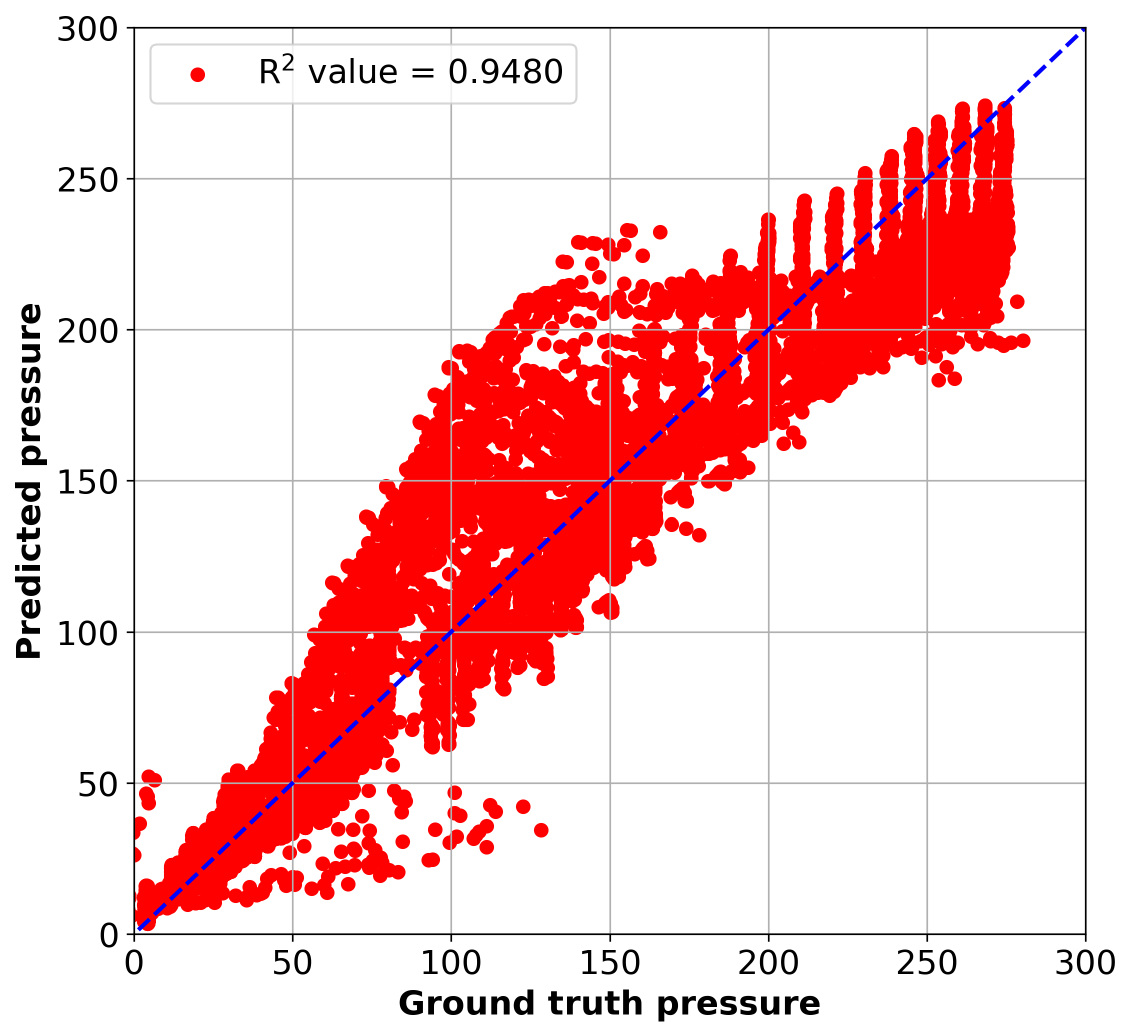}}
    \subfigure[]{
    \includegraphics[width=.35\textwidth]{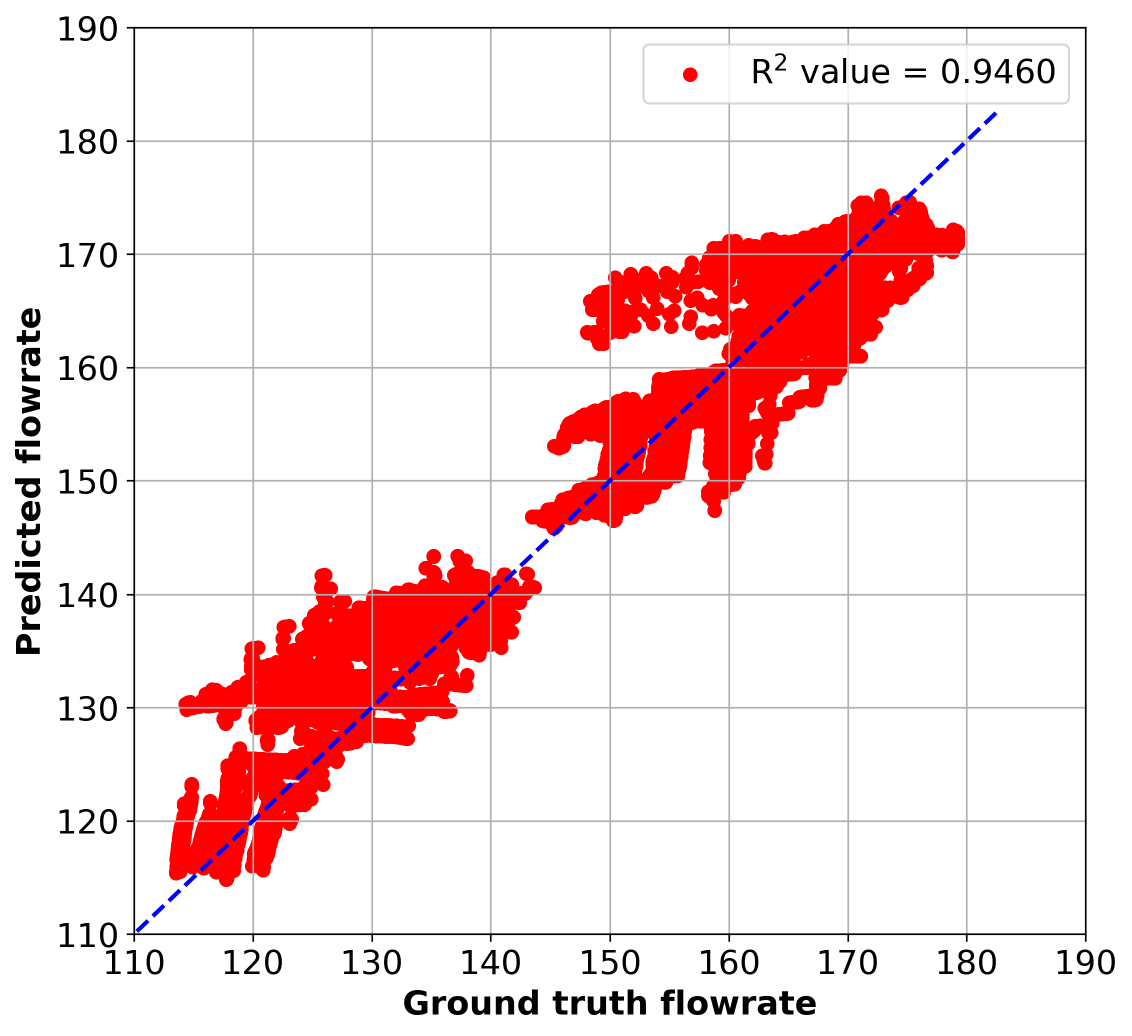}}
    \caption{Scatter plots showing the correlation between the predicted response and the ground truth for dataset 2:(a) Pressure, (b) Flow rate}
    \label{scatter2}
\end{figure} 

Prediction results are depicted in the \autoref{case2:pressure}, \autoref{case2:flowrate} and \autoref{scatter2}. The 3D visualizations in \autoref{case2:pressure}, \autoref{case2:flowrate} compare the predicted pressure fields and flow rates with the ground truth at various time steps. A close match between the results obtained using the proposed approach and the ground truth is observed.
The scatter plots of the pressure and flow rate shown in \autoref{scatter2} indicate a close match between the results obtained using the proposed approach and the ground truth. This is further indicated by the fact that the R-squared values are above $0.9$ ($0.9480$ and $0.9460$).
%
\begin{figure}[!ht]
    \centering{
    \includegraphics[width=0.6\textwidth]{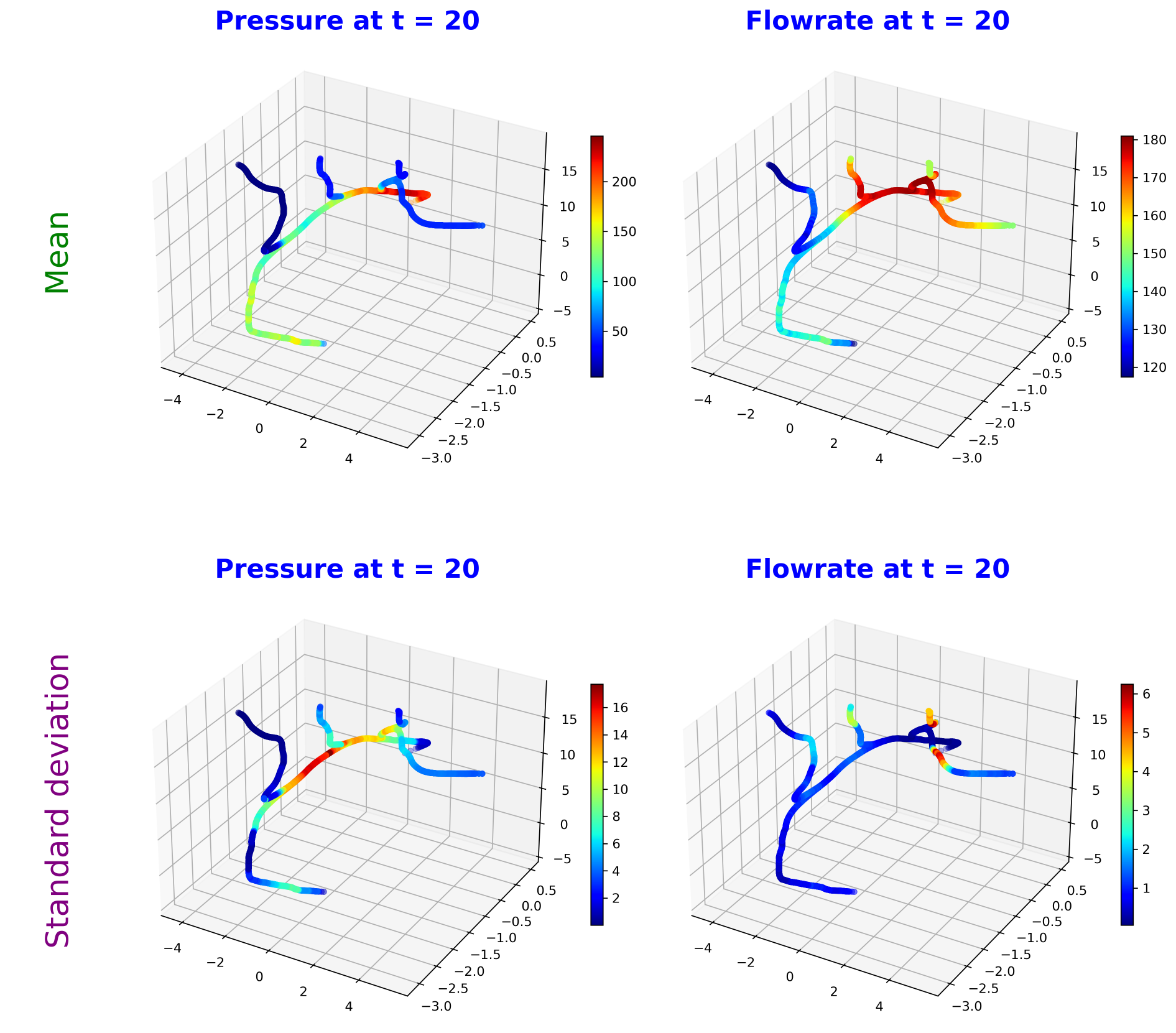}}
    \caption{Mean and standard deviation of the predicted responses of the 
    aorta model affected by coarctation under uncertainty in the initial conditions.}\label{case2:mean_std2}
\end{figure}

\begin{figure}[!ht]
    \centering{
    \includegraphics[width=0.6\textwidth]{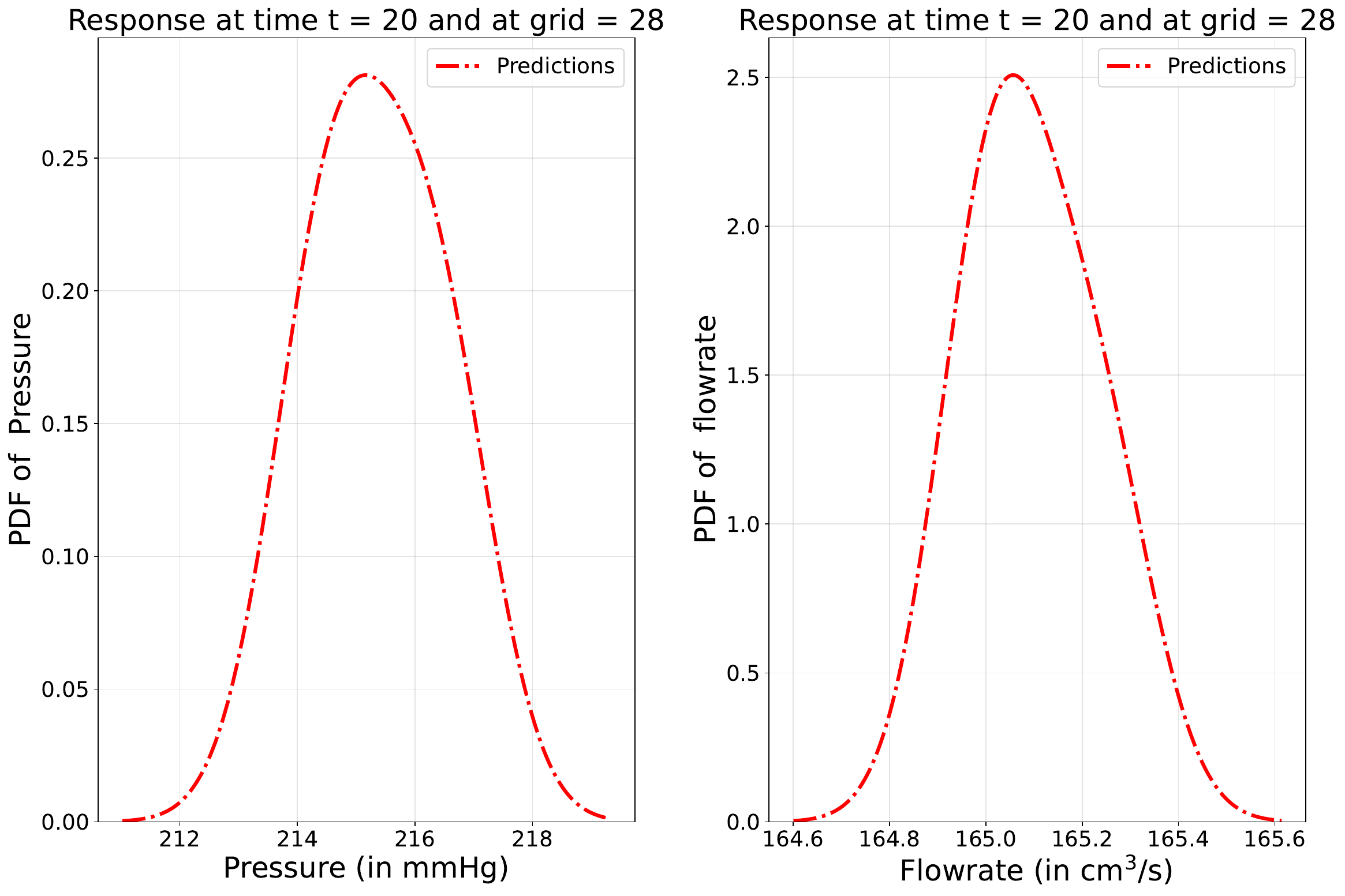}}
    \caption{Response PDF of the pressure and flowrate fields of the 
    aorta model affected by coarctation}\label{case2:pdf}
\end{figure}
Similar to the previous case study, the validated model is employed to quantify the propagation of the uncertainty from the input to the output. A similar setup (for the input uncertainty) as the previous case study is considered. The results are illustrated in \autoref{case2:mean_std2} and \autoref{case2:pdf}. \autoref{case2:mean_std2} showcase the mean and standard deviation of the fields at a specific time point, while \autoref{case2:pdf} showcase the probability density function of responses at a particular time and spatial location.
It can be seen in the mean pressure plot that blood pressure builds at upstream of the coarctation. The possible reason here is the narrowing of the of blood vessels which leads to a significant disruption of blood flow and increased resistance.  Similarly, the mean flow rate plot highlights a higher flow velocity at the location of coarctation, with subsequent flow disturbances downstream as the blood velocities adjust to the altered conditions. The standard deviation plot of pressure indicates significant variability at downstream of the coarctation due to dynamic pressure adjustments, whereas the standard deviation plot of the flow rate indicates high variability at the narrowing and in smaller downstream regions due to the turbulent flow dynamics caused by the constriction \cite{ncbi_coarctation}.

\subsection{Healthy aorta model 1} 
\begin{figure}[!ht]
    \centering{
    \includegraphics[width=1.0\textwidth]{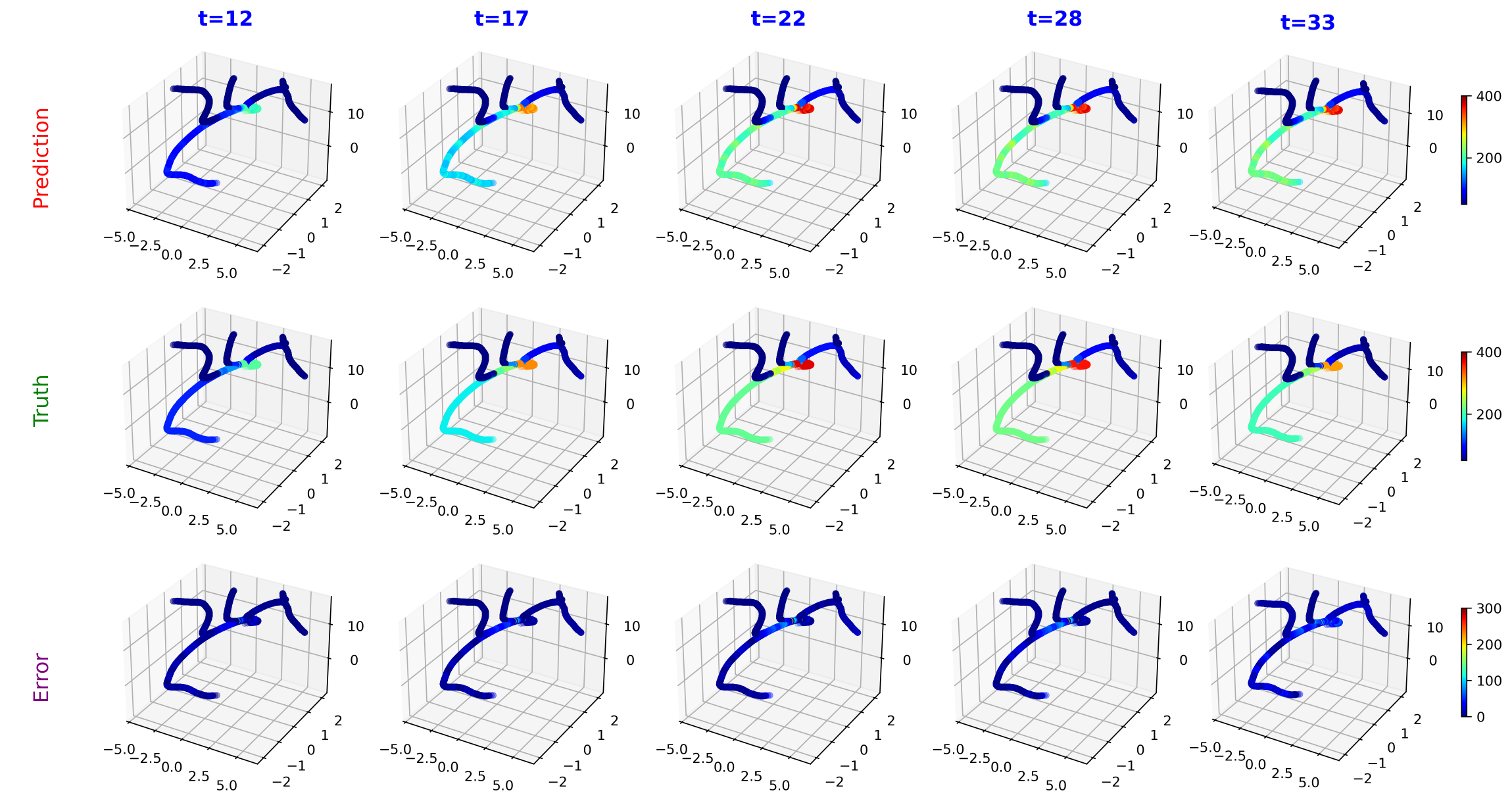}}
    \caption{Predictions results of pressure field measured in $mmHg$ for the dataset 3: (Top to bottom) Geometry adaptive waveformer prediction, ground truth response, and $L_1$ error.}\label{case3:pressure}
\end{figure}

\begin{figure}[!ht]
    \centering{
    \includegraphics[width=1.0\textwidth]{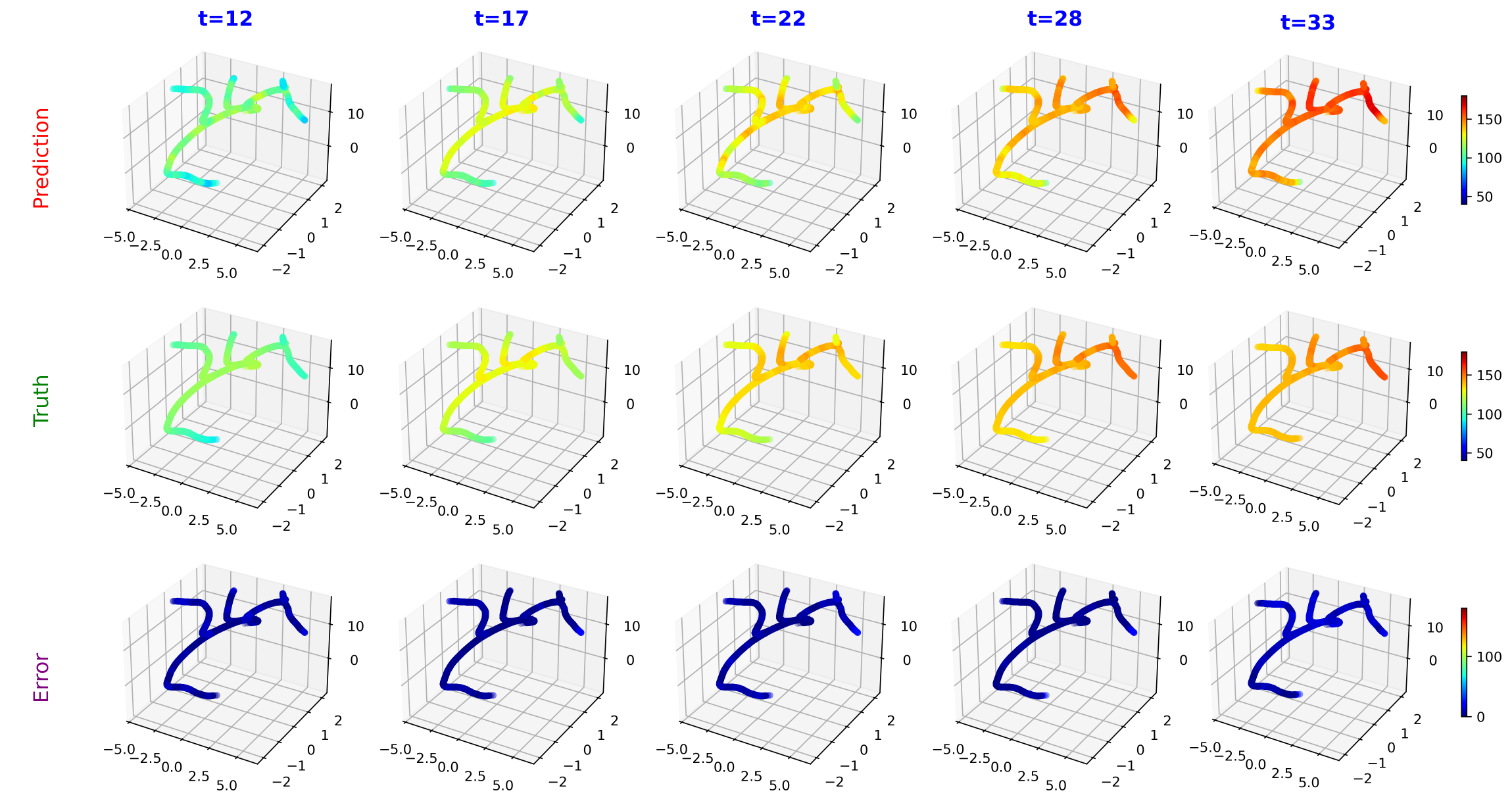}}
    \caption{Predictions results of flow rates measured in $cm^{3}/s$ for the dataset 2: (Top to bottom) Geometry adaptive waveformer prediction, ground truth response, and $L_1$ error.}\label{case3:flowrate}
\end{figure}

\begin{figure}[!ht]
    \centering
    \subfigure[]{
    \includegraphics[width=.35\textwidth]{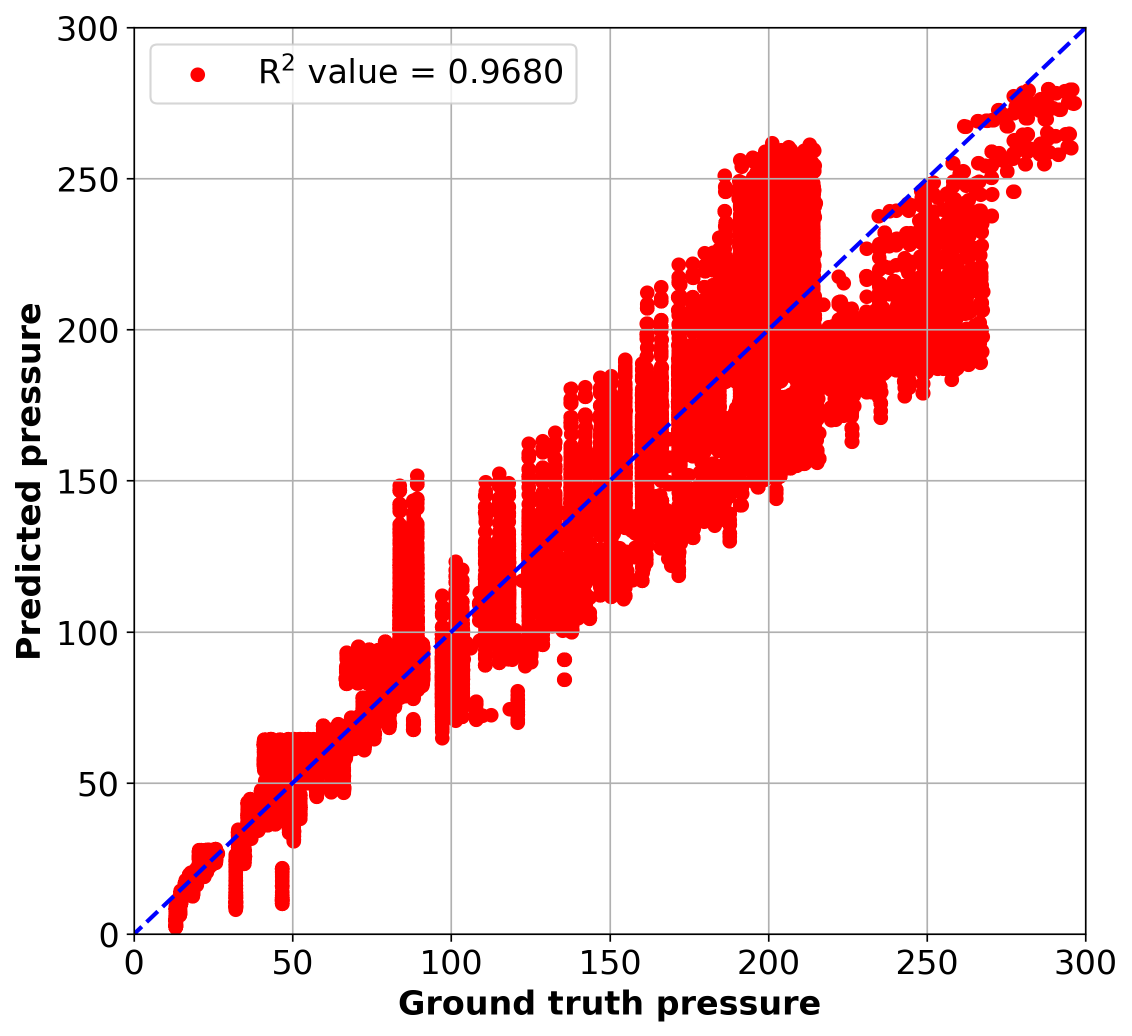}}
    \subfigure[]{
    \includegraphics[width=.35\textwidth]{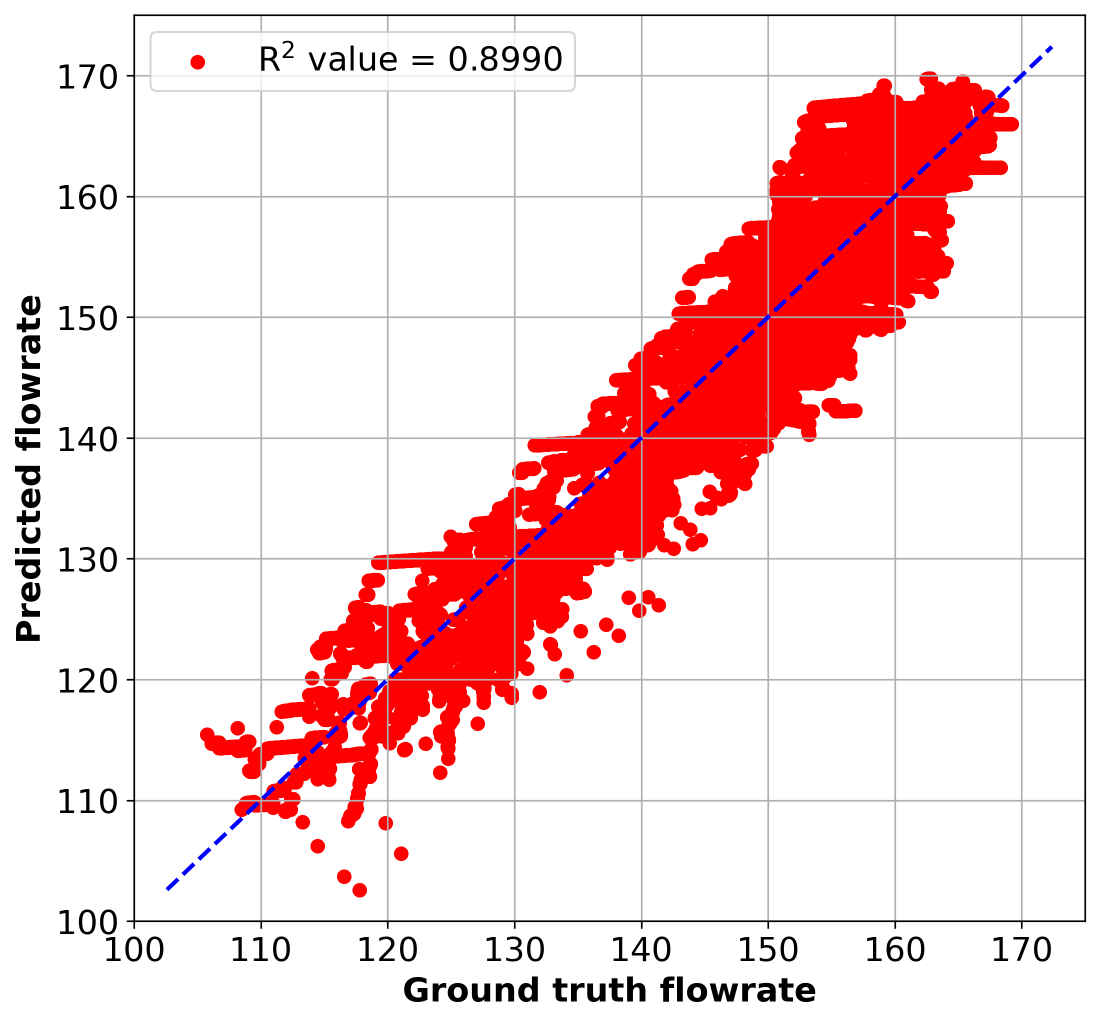}}
    \caption{Scatter plots showing the correlation between the predicted response and the ground truth for dataset 3:(a) Pressure, (b) Flow rate }
    \label{scatter3}
\end{figure}

For this case study, we consider cardiovascular flow corresponding to a healthy aorta model. The geometry adaptive waveformer is employed here to map spatiotemporal cardiovascular responses from an initial sequence of $k=10$ time steps to the subsequent $n=23$ time steps, where the responses are defined across 5102 discretized spatial locations. The training set has 27 trajectories, and the testing set contains 5 trajectories of vascular responses. The results obtained after training the proposed frameworks are presented in \autoref{case3:pressure}, \autoref{case3:flowrate}, and \autoref{scatter3}.
The visual depiction of the predicted results obtained for the pressure (shown in \autoref{case3:pressure}) and the flow rate (shown in \autoref{case3:flowrate}) evidently shows a close agreement with the ground truth. The scatter maps in \autoref{scatter3} indicate that the model performs reasonably well overall, though the R-squared value for the flow rate predictions falls slightly below $0.9$. This may be attributed to the fact that the model was trained with a relatively small number of samples. 

\begin{figure}[!ht]
    \centering{
    \includegraphics[width=0.6\textwidth]{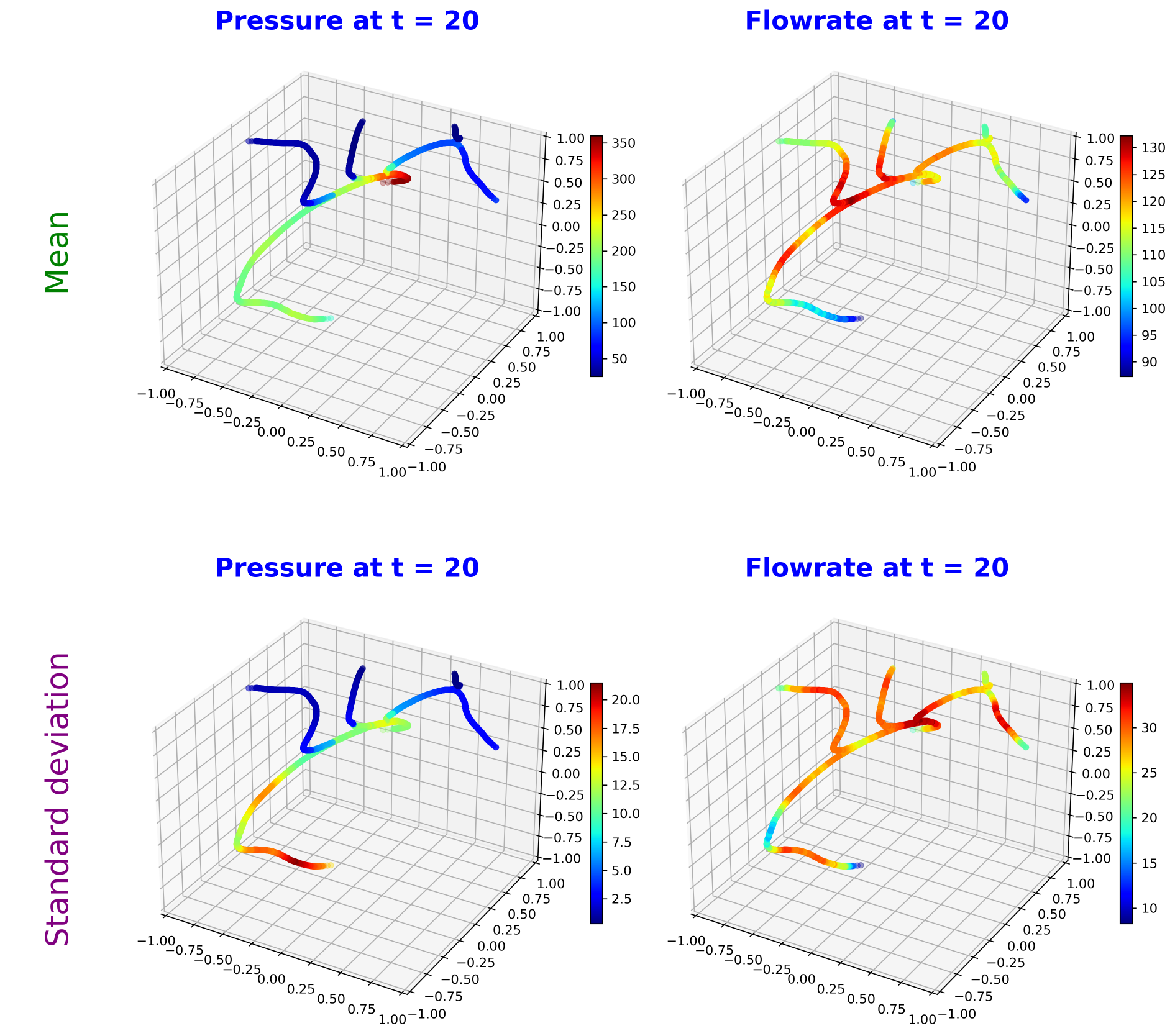}}
    \caption{Mean and standard deviation of the predicted responses of the 
    healthy aorta model 1 under uncertainty in the initial conditions.}\label{case3:mean_std3}
\end{figure}

\begin{figure}[!ht]
    \centering{
    \includegraphics[width=0.6\textwidth]{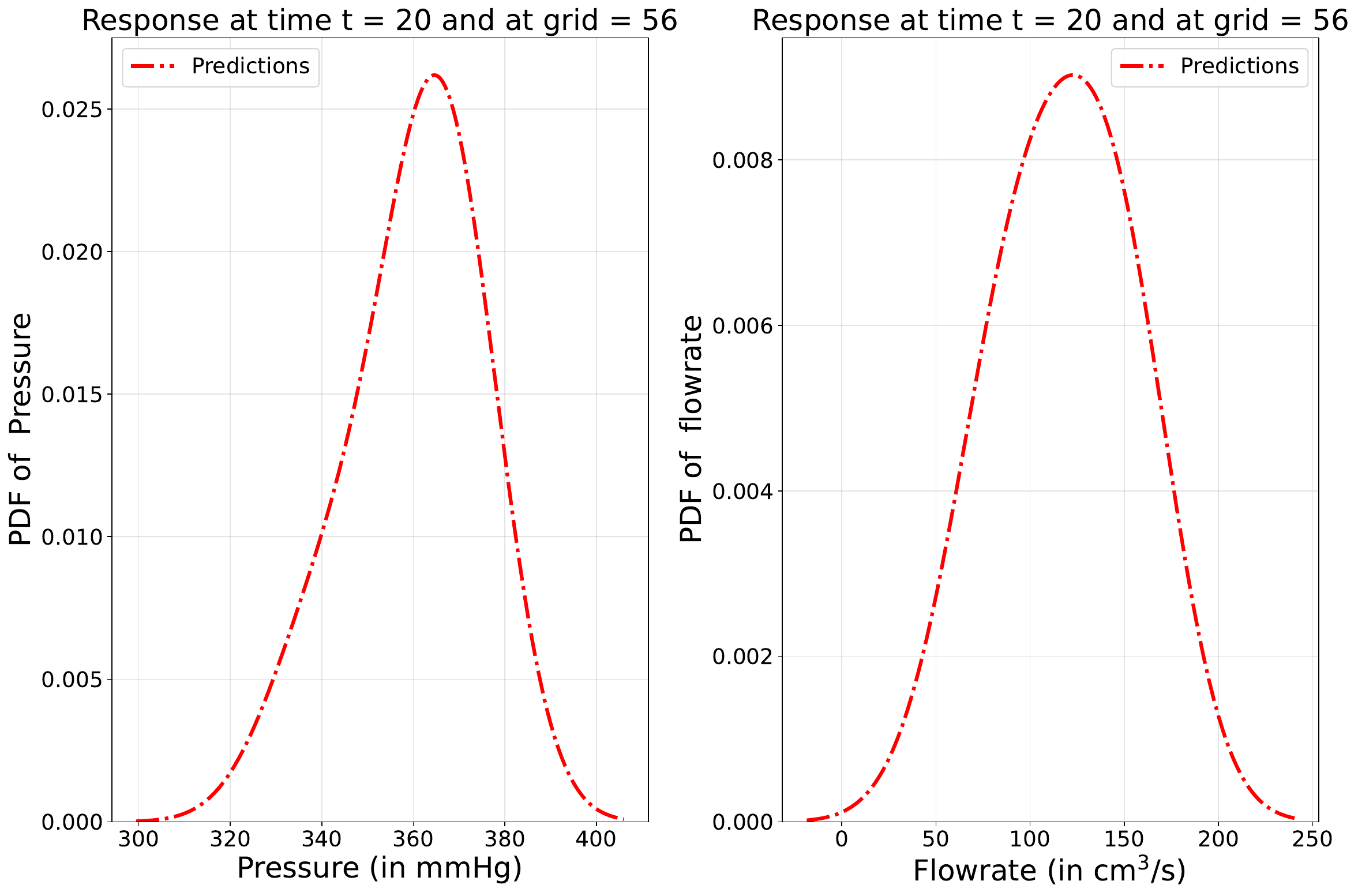}}
    \caption{Response PDF of the pressure and flowrate fields of the healthy aorta model 1}\label{case3:pdf}
\end{figure}


Next, the geometry adaptive waveformer is employed to quantify the propagated uncertainty in the response variables. Uncertainty in the initial condition is considered similar to the previous cases. 
\autoref{case3:mean_std3} illustrates the mean and standard deviation of the flow fields obtained using the proposed approach. Additionally, \autoref{case3:pdf} provides the probability density function of responses at a specific time and spatial location. It can be observed from the mean plots that a gradual decline in pressure along the length of the aorta starts from higher values near the inlet. The smooth transition to a lower pressure downstream indicates unrestricted blood flow and minimal flow resistance within the vessel. Similarly, the mean flow rate highlights high flow velocities in the central portion of the aorta, where the main arterial pathway naturally carries most of the blood. Flow velocities decrease smoothly downstream and near the branching regions as blood splits into smaller vessels. In regards to the variability of the flow, the pressure standard deviation plot indicates low variability across most regions of the aorta, with slightly higher values near the inlet and along curvatures. This marginal variability is expected due to flow adjustments at the inlet and natural geometric changes. The flow rate standard deviation plot shows higher variability at the main branching points, where the blood flow splits into smaller vessels. This localized variability is a result of flow redistribution and minor asymmetries during branching, which are normal for a healthy aorta. Thus, overall the flow patterns indicate a consistent, stable, laminar blood flow in a healthy aorta without significant disturbances.

\subsection{Healthy aorta model 2} 
Lastly, we consider another healthy aorta model; however, with a different geometric configuration.
For this model, the responses are defined on 
5183 non-uniformly discretized spatial points.
Similar to the previous cases, we consider the responses corresponding to the first $k=10$ time-steps as input and the next
$n=15$ steps as output. 
The training and test data sets are comprised of 27 and 5 trajectories, respectively. Other setups remain same as the previous examples.
The pressure and flow fields obtained at different time-instants are shown in \autoref{case3:pressure}, \autoref{case3:flowrate}, and \autoref{scatter3}.
\begin{figure}[!ht]
    \centering{
    \includegraphics[width=1.0\textwidth]{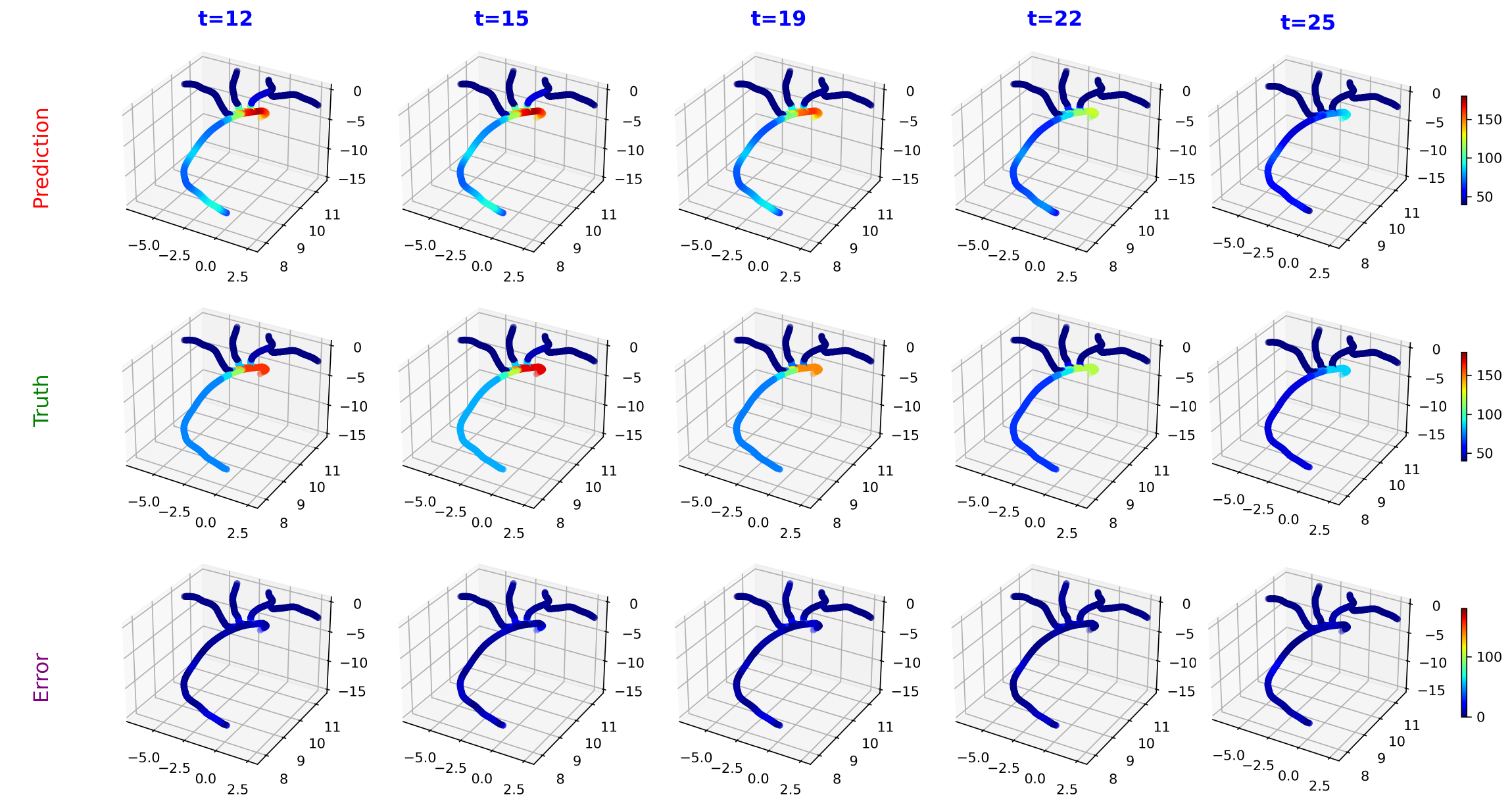}}
    \caption{Predictions results of pressure field measured in $mmHg$ for the dataset 4: (Top to bottom) Geometry adaptive waveformer prediction, ground truth response, and $L_1$ error.}\label{case4:pressure}
\end{figure}
\begin{figure}[!ht]
    \centering{
    \includegraphics[width=1.0\textwidth]{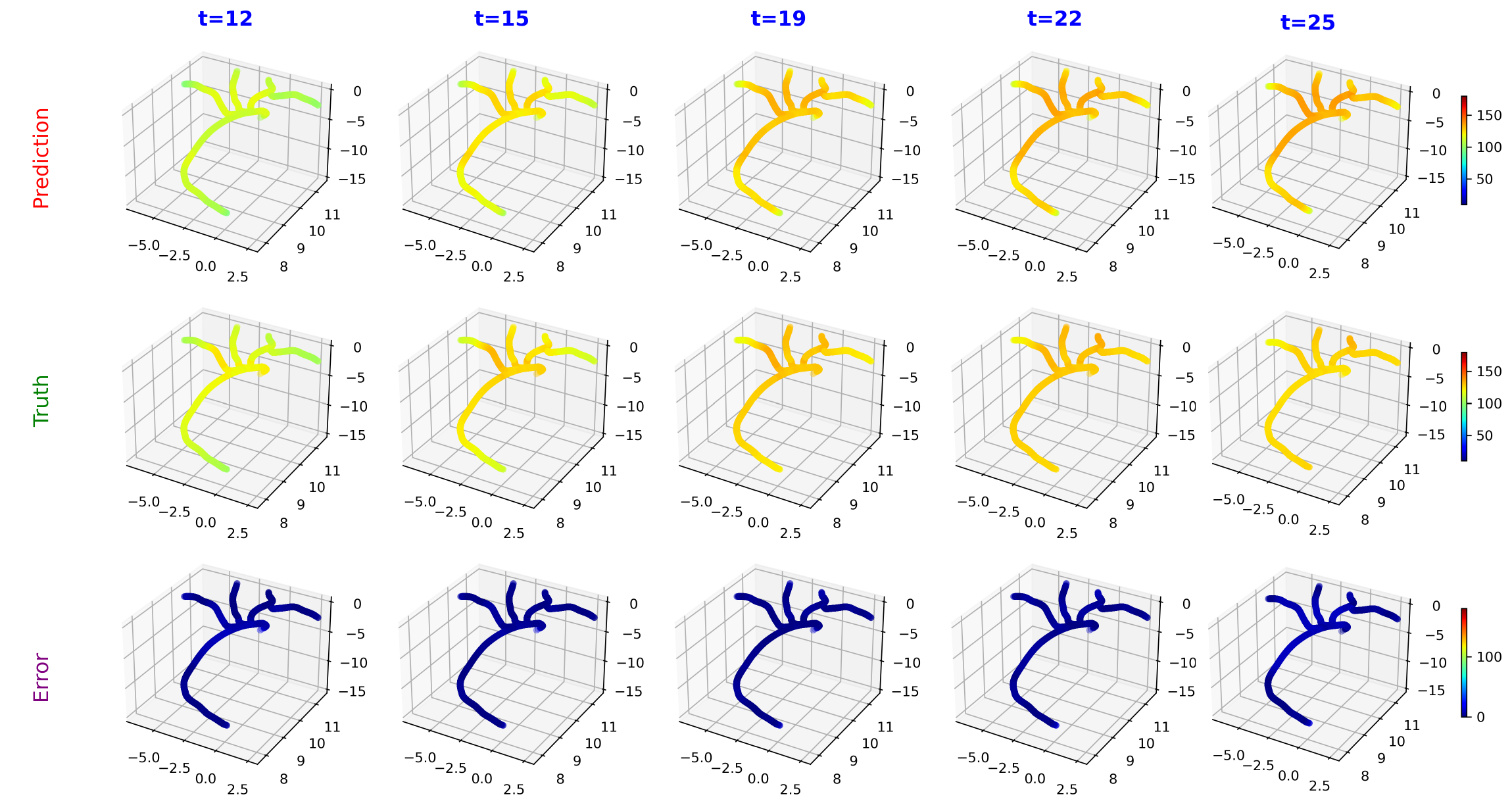}}
    \caption{Predictions results of flow rates measured in $cm^{3}/s$ for the dataset 2: (Top to bottom) Geometry adaptive waveformer prediction, ground truth response, and $L_1$ error.}\label{case4:flowrate}
\end{figure}

\begin{figure}[!ht]
    \centering
    \subfigure[]{
    \includegraphics[width=.35\textwidth]{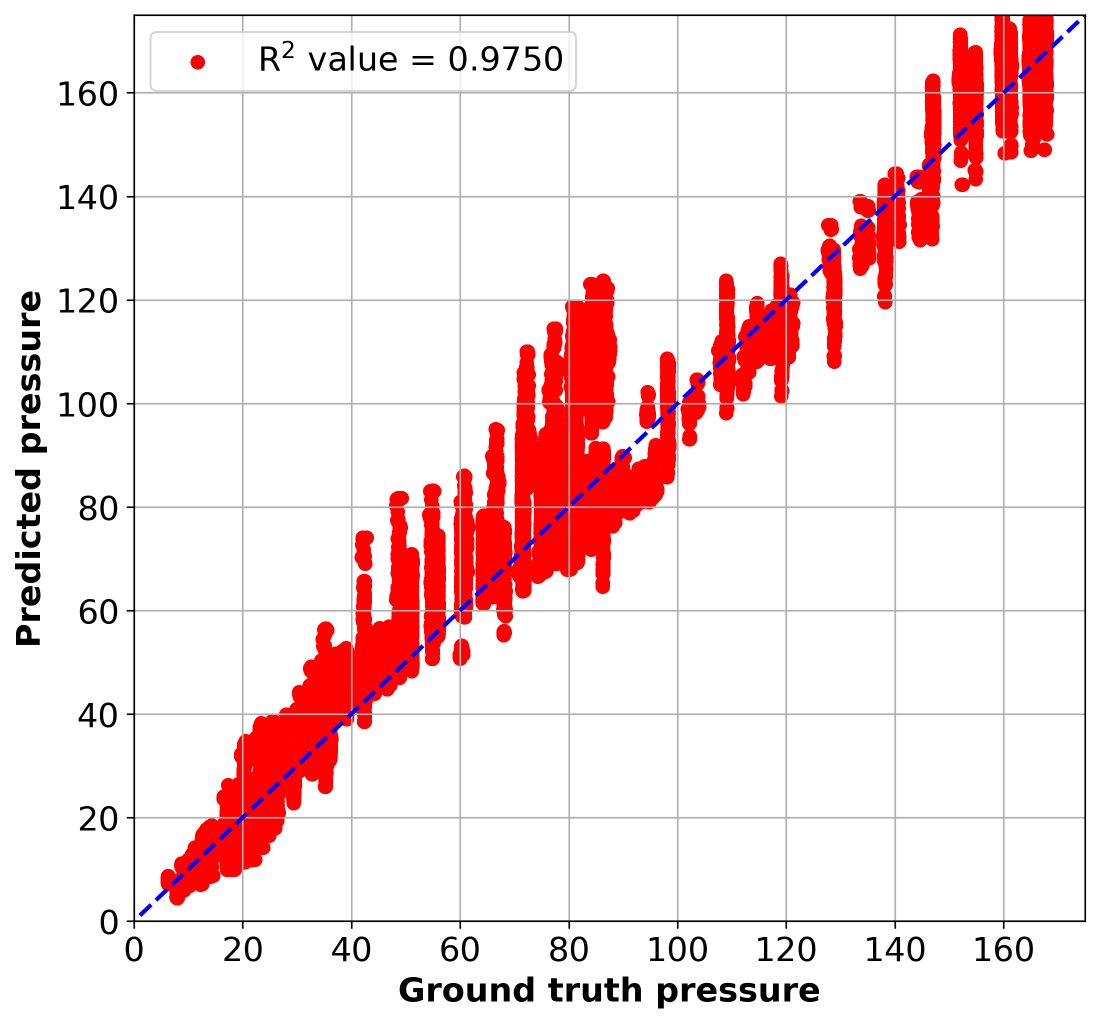}}
    \subfigure[]{
    \includegraphics[width=.35\textwidth]{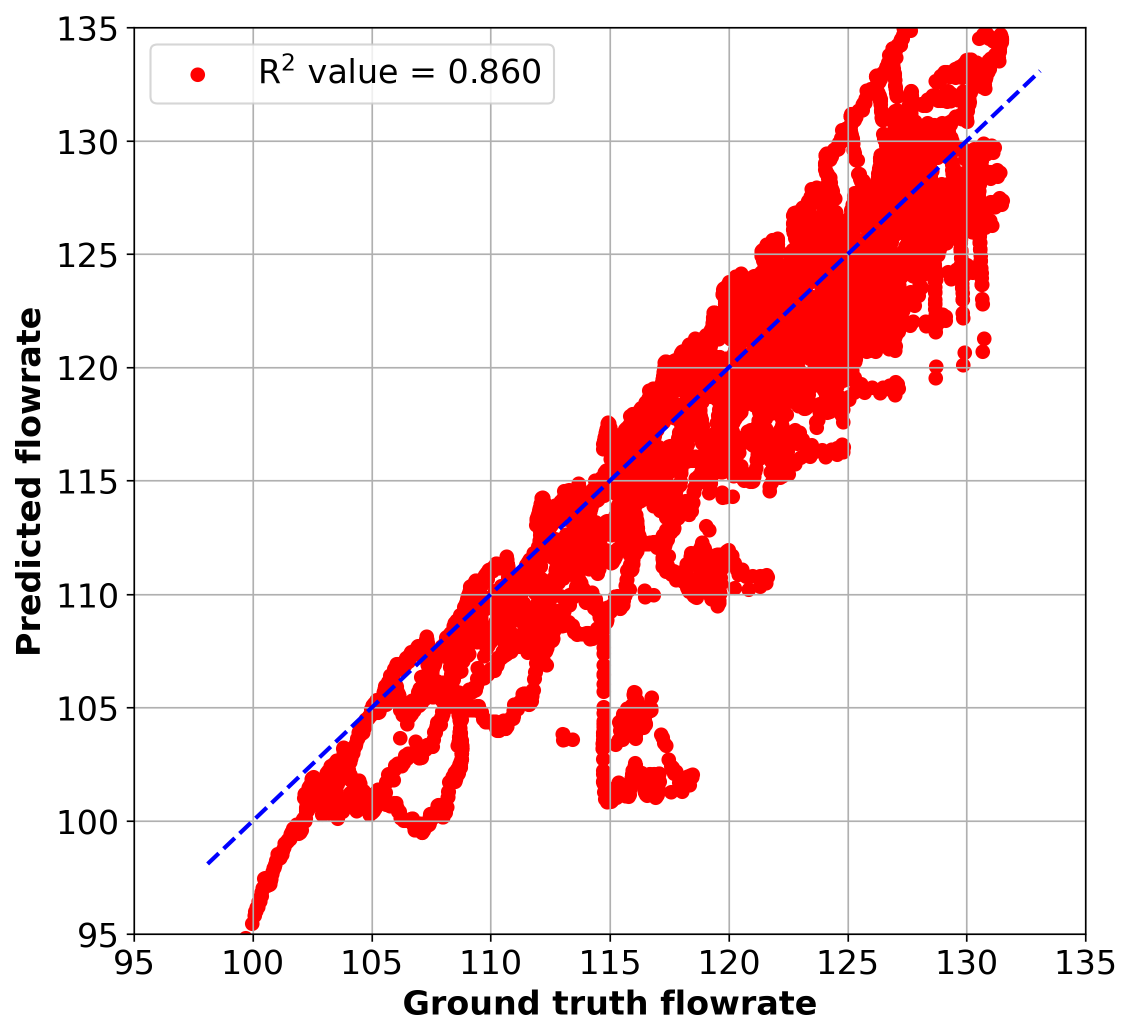}}
    \caption{Scatter plots showing the correlation between the predicted response and the ground truth for dataset 4:(a) Pressure, (b) Flow rate }
    \label{scatter4}
\end{figure} 
Visually, it is evident that the predictions match well with the ground truth of vascular responses across different time steps. Moreover, the scatter plot provided in \autoref{scatter4} shows a good match between the results obtained using the proposed approach and the ground truth. 

\begin{figure}[!ht]
    \centering{
    \includegraphics[width=0.6\textwidth]{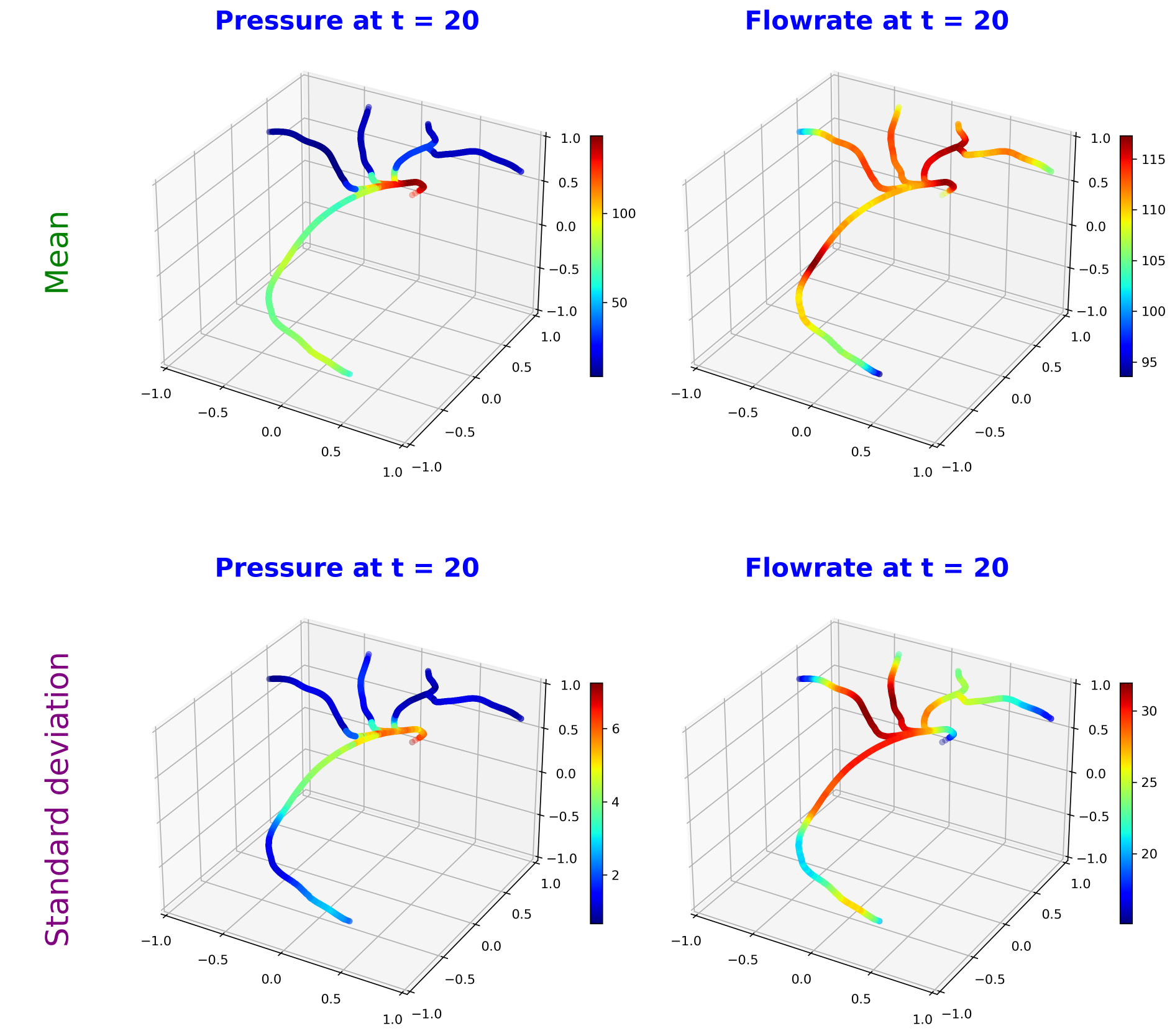}}
    \caption{Mean and standard deviation of the predicted responses of the 
    healthy aorta model 2 under uncertainty in the initial conditions.}\label{case4:mean_std4}
\end{figure}

\begin{figure}[!ht]
    \centering{
    \includegraphics[width=0.6\textwidth]{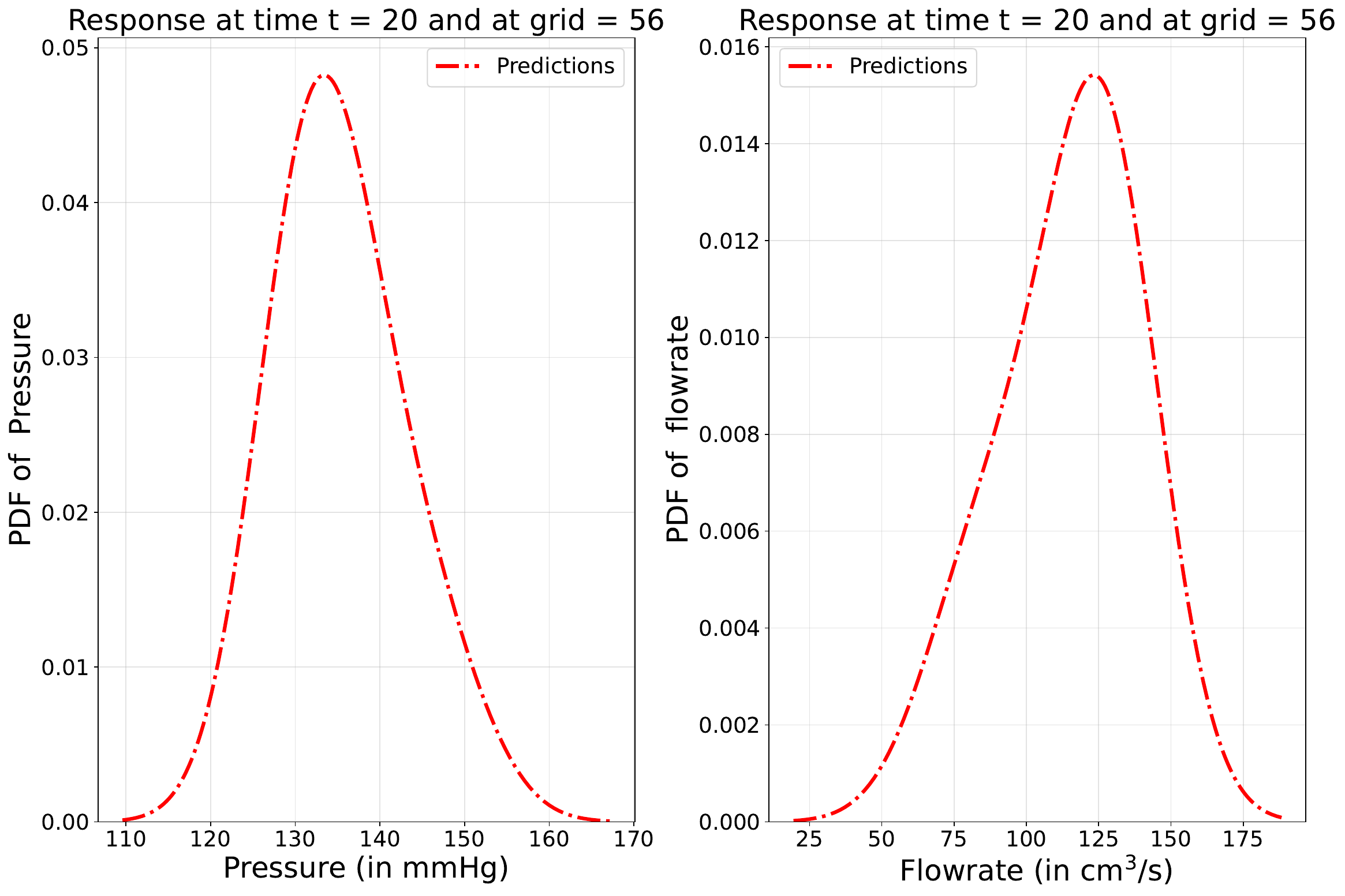}}
    \caption{Response PDF of the pressure and flowrate fields of the healthy aorta model 2}\label{case4:pdf}
\end{figure}

To estimate the propagated uncertainty in the fields resulting from perturbations in the initial conditions, the trained model is employed to predict time-marched responses. The results of the study are illustrated in \autoref{case4:mean_std4}, which presents the mean and standard deviation of the pressure and flow rate fields at a specific time instants ($t=20$). \autoref{case4:pdf}, on the other hand, showcases the probability density of the responses at designated times and spatial locations.
It can be observed from the results that similar to the previous case of the healthy aorta model, the mean pressure plot exhibits a gradual decrease along the aorta, starting from higher values at the inlet and smoothly transitioning to lower values downstream. The mean flow rate plot highlights higher velocities along the main arterial pathway, particularly near the centerline and inlet regions. As the blood progresses downstream and branches into smaller vessels, the flow rate gradually decreases. Overall, the mean response plots provide a clear indication of a healthy aortic structure that allows for efficient pressure dissipation, consistent flow distribution, and laminar flow behavior. Further analysis of the standard deviation plots for pressure and flow rate adds better insights. The pressure standard deviation plot shows low variability across the vessel, with slightly higher values near the inlet and branching points due to natural flow adjustments. Similarly, the flow rate standard deviation plot indicates higher variability near the branching points and along the main arterial pathway. This localized variability arises from flow splitting and minor adjustments as blood divides into smaller branches. Overall, the plots reinforce the natural flow behavior, smooth pressure drop, and redistribution of blood flow at bifurcation points, which is consistent with a healthy aorta.

\subsection{Prediction over entire cardiac cycle}
\begin{figure}[!ht]
    \centering
    \subfigure[]{
    \includegraphics[width=0.45\textwidth]{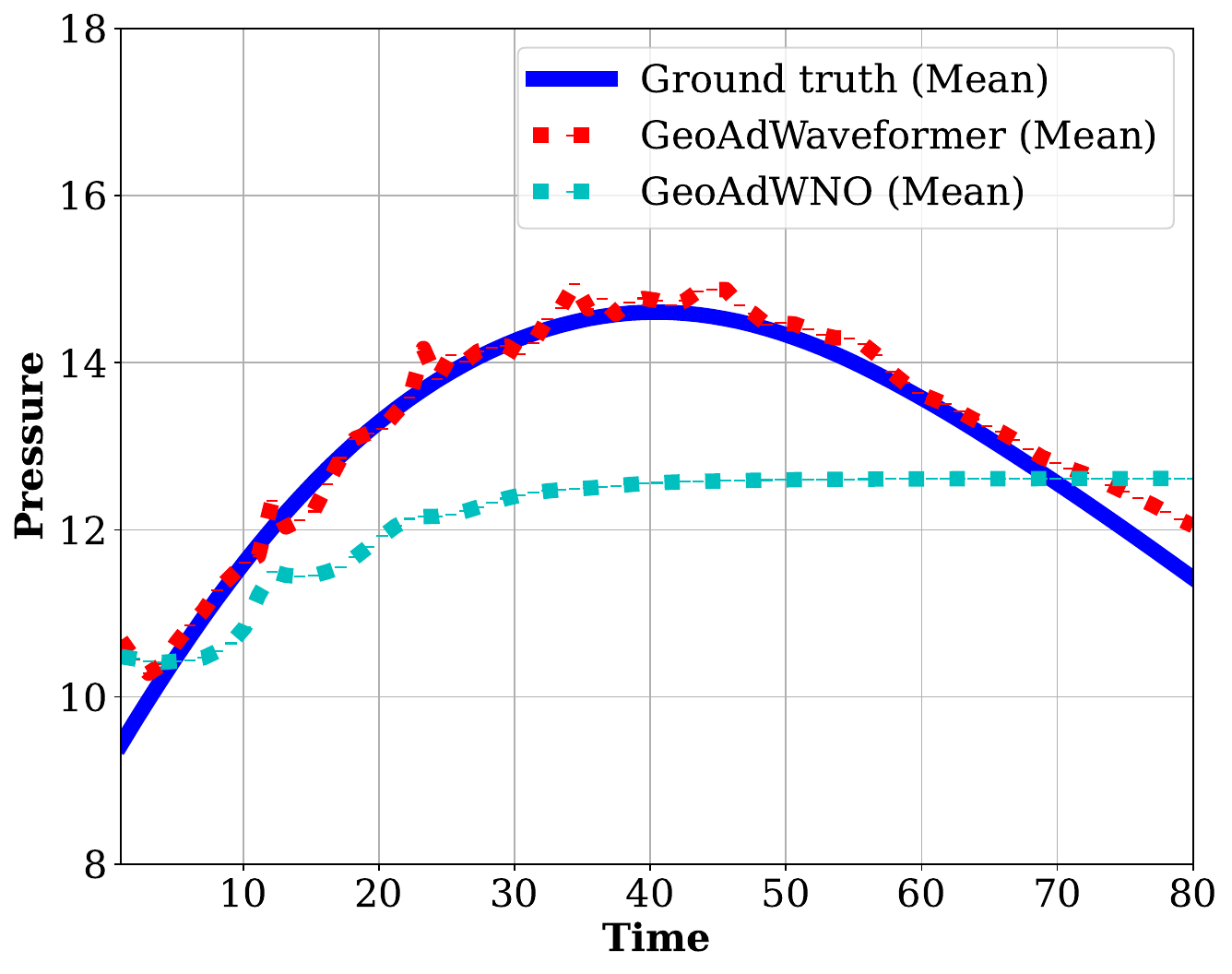}}
    \subfigure[]{
    \includegraphics[width=0.45\textwidth]{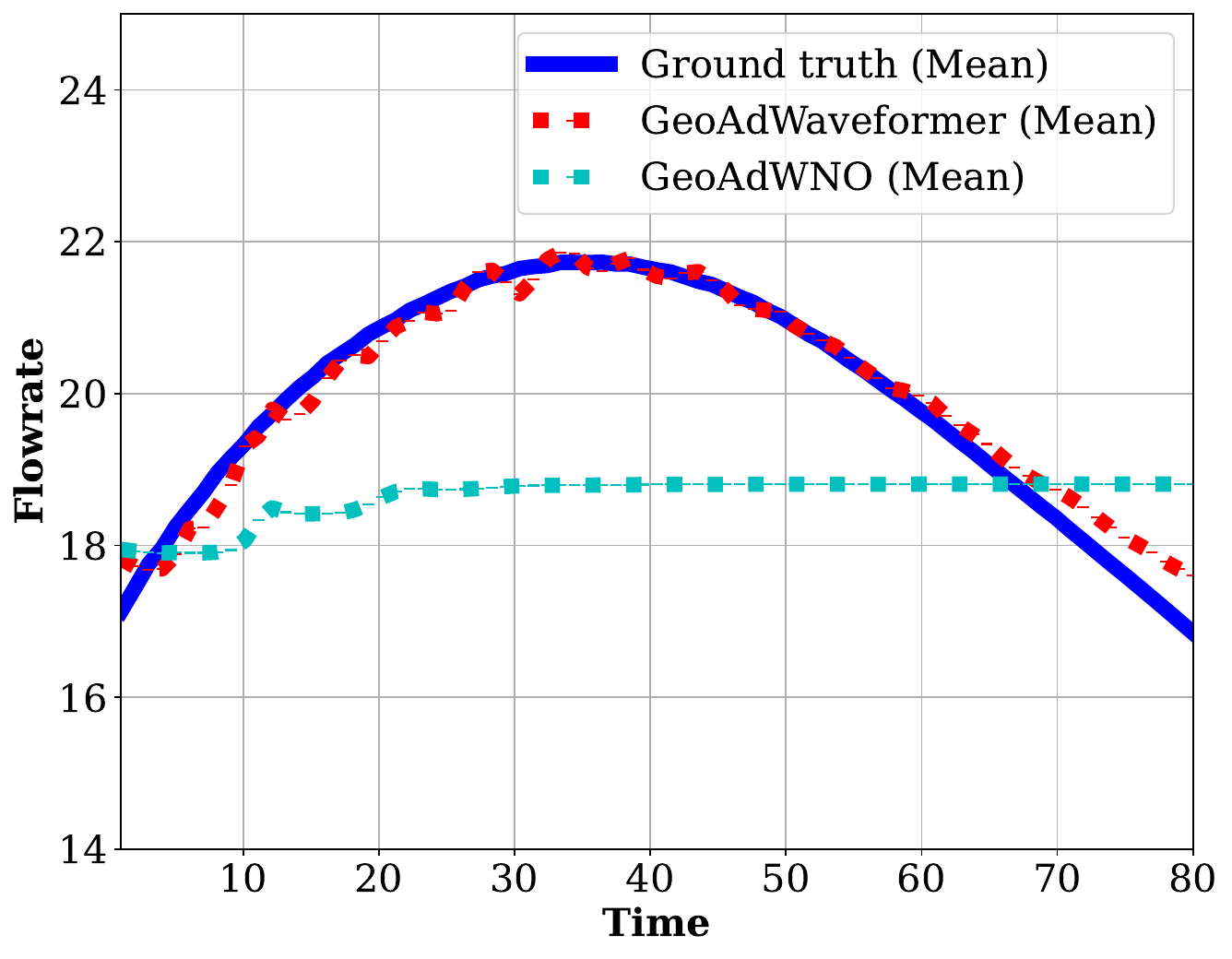}}
    \caption{Variation of the mean responses with time :(a) Pressure, (b) Flow rate, where the mean of the ground truth, predictions of geometry adaptive waveformer, and predictions of the geometry adaptive WNO are plotted against time}
    \label{Respose_comp}
\end{figure}

An additional study is conducted to obtain the response prediction for the entire cardiac cycle. Specifically, we strive to obtain a prediction of the entire cycle only from the first initial state, extending through to the end time of the cycle. A modified geometry-adaptive waveformer is employed here to enhance training efficiency. The details of the training scheme and the modified architecture are provided in \autoref{fig:modifiedgwaveformer} (see Remark in Section \ref{subsec:geom_adap_wf}). The results of the study are presented in \autoref{Respose_comp}, which demonstrates variations of mean responses of geometry adaptive waveformer in comparison with that of geometry adaptive WNO predictions and ground truths over the one cardiac cycle. We observe that the results obtained using the proposed approach outperform WNO and match well with the ground truth data.

\section{Conclusions}\label{sec:conclusion}
In this work, we propose a geometry-adaptive waveformer for modeling cardiovascular dynamics. Our approach aims to enhance the workflow of patient-specific cardiovascular analysis by mitigating the computational bottleneck associated with patient-specific simulations. The prime challenge in simulating the vascular structure stems from the complex and irregular structures of the cardiovascular anatomies. While patient-specific numerical simulations are computationally expensive, effective data-driven modeling is often not feasible as cardiovascular modeling involves dealing with a high-dimensional, time-dependent problem defined on an irregular domain. To that end, we propose a geometry-adaptive waveformer to address this challenge by transforming inputs defined on irregular geometries into a regular latent space, where the dynamics are effectively learned. The predictions are then mapped back to the original domain, offering a robust and efficient solution for cardiovascular modeling.

We rigorously validate the performance of the geometry adaptive waveformer using diverse cardiovascular datasets, including the simulated responses of the pulmonary artery, healthy aorta, and diseased aortas. These datasets encompass a wide range of vascular dynamics, characterized by irregular geometries and complex spatiotemporal behaviors. By effectively mapping the cardiovascular responses from an initial sequence of time steps to future time steps, the waveformer demonstrates its capacity to address the computational challenges inherent to patient-specific cardiovascular modeling. The results demonstrate the efficacy of the proposed framework in accurately predicting vascular responses, highlighting its potential application in propagating uncertainty in input to the response. Furthermore, the geometry-adaptive nature of the waveformer allows it to transform irregular spatial domains into a regular latent space, enabling efficient learning of the dynamics and robust mapping back to the original domain. This capability not only ensures accuracy but also reinforces the potential of the proposed framework to enhance computational efficiency in cardiovascular simulations significantly. The key findings of the study are outlined below:
\begin{itemize}
    \item \textbf{Geometry adaptation :} The geometry-adaptive waveformer efficiently maps inputs from irregular spatial domains to a structured latent space, enabling effective learning of high-dimensional dynamics and precise reconstruction in the original domain.
    \item \textbf{Temporal learning :} The waveformer effectively captures long-term temporal behaviors.
    \item \textbf{Generalizability:} The framework exhibits strong adaptability to diverse 3D geometries and high-dimensional systems, demonstrating its versatility for various complex problems.
    \item \textbf{Uncertainty propagation:} The waveformer facilitates robust uncertainty propagation, ensuring accurate and reliable predictions under varying input conditions.
\end{itemize}

\section*{Acknowledgements}
NN acknowledges the support received from the Ministry of Education in the form of the Prime Minister's Research Fellowship. SC acknowledges the financial support received from the Ministry of Ports and Shipping via letter number ST-14011/74/MT (356529) and the Science and Engineering Research Board via grant number CRG/2023/007667


\begin{thebibliography}{10}

\bibitem{gaziano2006cardiovascular}
T.~Gaziano, K.~S. Reddy, F.~Paccaud, S.~Horton, V.~Chaturvedi, Cardiovascular disease, Disease Control Priorities in Developing Countries. 2nd edition (2006).

\bibitem{bao2014usnctam}
G.~Bao, Y.~Bazilevs, J.-H. Chung, P.~Decuzzi, H.~D. Espinosa, M.~Ferrari, H.~Gao, S.~S. Hossain, T.~J. Hughes, R.~D. Kamm, et~al., Usnctam perspectives on mechanics in medicine, Journal of The Royal Society Interface 11~(97) (2014) 20140301.

\bibitem{schwarz2023beyond}
E.~L. Schwarz, L.~Pegolotti, M.~R. Pfaller, A.~L. Marsden, Beyond cfd: Emerging methodologies for predictive simulation in cardiovascular health and disease, Biophysics Reviews 4~(1) (2023).

\bibitem{zhong2018application}
L.~Zhong, J.-M. Zhang, B.~Su, R.~S. Tan, J.~C. Allen, G.~S. Kassab, Application of patient-specific computational fluid dynamics in coronary and intra-cardiac flow simulations: Challenges and opportunities, Frontiers in physiology 9 (2018) 742.

\bibitem{stuhne2004finite}
G.~R. Stuhne~and, D.~A. Steinman, Finite-element modeling of the hemodynamics of stented aneurysms, J. Biomech. Eng. 126~(3) (2004) 382--387.

\bibitem{eslami2020effect}
P.~Eslami, J.~Tran, Z.~Jin, J.~Karady, R.~Sotoodeh, M.~T. Lu, U.~Hoffmann, A.~Marsden, Effect of wall elasticity on hemodynamics and wall shear stress in patient-specific simulations in the coronary arteries, Journal of biomechanical engineering 142~(2) (2020) 024503.

\bibitem{syed2023modeling}
F.~Syed, S.~Khan, M.~Toma, Modeling dynamics of the cardiovascular system using fluid-structure interaction methods, Biology 12~(7) (2023) 1026.

\bibitem{crosetto2011fluid}
P.~Crosetto, P.~Reymond, S.~Deparis, D.~Kontaxakis, N.~Stergiopulos, A.~Quarteroni, Fluid--structure interaction simulation of aortic blood flow, Computers \& Fluids 43~(1) (2011) 46--57.

\bibitem{tay2011towards}
W.-B. Tay, Y.-H. Tseng, L.-Y. Lin, W.-Y. Tseng, Towards patient-specific cardiovascular modeling system using the immersed boundary technique, Biomedical engineering online 10 (2011) 1--17.

\bibitem{sarkar2017immersed}
D.~Sarkar, N.~Upadhyay, S.~Roy, S.~C. Rana, Immersed boundary simulation of flow through arterial junctions, S{\=a}dhan{\=a} 42 (2017) 533--541.

\bibitem{pontrelli2014lattice}
G.~Pontrelli, I.~Halliday, S.~Melchionna, T.~J. Spencer, S.~Succi, Lattice boltzmann method as a computational framework for multiscale haemodynamics, Mathematical and Computer Modelling of Dynamical Systems 20~(5) (2014) 470--490.

\bibitem{dal2020reduced}
N.~Dal~Santo, A.~Manzoni, S.~Pagani, A.~Quarteroni, Reduced-order modeling for applications to the cardiovascular system, Applications; De Gruyter: Berlin, Germany (2020) 251--278.

\bibitem{zhang2020personalized}
X.~Zhang, D.~Wu, F.~Miao, H.~Liu, Y.~Li, Personalized hemodynamic modeling of the human cardiovascular system: a reduced-order computing model, IEEE Transactions on Biomedical Engineering 67~(10) (2020) 2754--2764.

\bibitem{gul2016mathematical}
R.~Gul, Mathematical modeling and sensitivity analysis of lumped-parameter model of the human cardiovascular system, Ph.D. thesis (2016).

\bibitem{liang2005closed}
F.~Liang, H.~Liu, A closed-loop lumped parameter computational model for human cardiovascular system, JSME International Journal Series C Mechanical Systems, Machine Elements and Manufacturing 48~(4) (2005) 484--493.

\bibitem{li2019method}
B.~Li, W.~Wang, B.~Mao, Y.~Liu, A method to personalize the lumped parameter model of coronary artery, International Journal of Computational Methods 16~(03) (2019) 1842004.

\bibitem{bandola2016identification}
D.~Bando{\l}a, Identification and modeling the pulsatile blood flow in the cardiovascular system using a zero-dimensional model in an electrical analogy, Archiwum Instytutu Techniki Cieplnej 1 (2016).

\bibitem{formaggia2003one}
L.~Formaggia, D.~Lamponi, A.~Quarteroni, One-dimensional models for blood flow in arteries, Journal of engineering mathematics 47 (2003) 251--276.

\bibitem{soudah2014reduced}
E.~Soudah, R.~Rossi, S.~Idelsohn, E.~O{\~n}ate, A reduced-order model based on the coupled 1d-3d finite element simulations for an efficient analysis of hemodynamics problems, Computational Mechanics 54 (2014) 1013--1022.

\bibitem{branen2021data}
A.~L. Branen, Data-driven modeling and control of cardiac system, Master's thesis, University of Idaho (2021).

\bibitem{ye2024data}
D.~Ye, V.~Krzhizhanovskaya, A.~G. Hoekstra, Data-driven reduced-order modelling for blood flow simulations with geometry-informed snapshots, Journal of Computational Physics 497 (2024) 112639.

\bibitem{siena2023data}
P.~Siena, M.~Girfoglio, F.~Ballarin, G.~Rozza, Data-driven reduced order modelling for patient-specific hemodynamics of coronary artery bypass grafts with physical and geometrical parameters, Journal of Scientific Computing 94~(2) (2023) 38.

\bibitem{pfaller2024reduced}
M.~R. Pfaller, L.~Pegolotti, J.~Pham, N.~L. Rubio, A.~L. Marsden, Reduced-order modeling of cardiovascular hemodynamics, in: Biomechanics of the Aorta, Elsevier, 2024, pp. 449--476.

\bibitem{sahli2020physics}
F.~Sahli~Costabal, Y.~Yang, P.~Perdikaris, D.~E. Hurtado, E.~Kuhl, Physics-informed neural networks for cardiac activation mapping, Frontiers in Physics 8 (2020) 42.

\bibitem{alzhanov2024three}
N.~Alzhanov, E.~Y. Ng, Y.~Zhao, Three-dimensional physics-informed neural network simulation in coronary artery trees, Fluids 9~(7) (2024) 153.

\bibitem{anandkumar2020neural}
A.~Anandkumar, K.~Azizzadenesheli, K.~Bhattacharya, N.~Kovachki, Z.~Li, B.~Liu, A.~Stuart, Neural operator: Graph kernel network for partial differential equations, in: ICLR 2020 Workshop on Integration of Deep Neural Models and Differential Equations, 2020.

\bibitem{li2020multipole}
Z.~Li, N.~Kovachki, K.~Azizzadenesheli, B.~Liu, A.~Stuart, K.~Bhattacharya, A.~Anandkumar, Multipole graph neural operator for parametric partial differential equations, Advances in Neural Information Processing Systems 33 (2020) 6755--6766.

\bibitem{cao2024laplace}
Q.~Cao, S.~Goswami, G.~E. Karniadakis, Laplace neural operator for solving differential equations, Nature Machine Intelligence (2024) 1--10.

\bibitem{li2020fourier}
Z.~Li, N.~Kovachki, K.~Azizzadenesheli, B.~Liu, K.~Bhattacharya, A.~Stuart, A.~Anandkumar, Fourier neural operator for parametric partial differential equations, arXiv preprint arXiv:2010.08895 (2020).

\bibitem{kovachki2021universal}
N.~Kovachki, S.~Lanthaler, S.~Mishra, On universal approximation and error bounds for fourier neural operators, The Journal of Machine Learning Research 22~(1) (2021) 13237--13312.

\bibitem{tripura2022wavelet}
T.~Tripura, S.~Chakraborty, Wavelet neural operator: a neural operator for parametric partial differential equations, arXiv preprint arXiv:2205.02191 (2022).

\bibitem{navaneeth2023physics}
N.~Navaneeth, T.~Tripura, S.~Chakraborty, Physics informed wno, arXiv preprint arXiv:2302.05925 (2023).

\bibitem{zhang2019wavelet}
D.~Zhang, D.~Zhang, Wavelet transform, Fundamentals of image data mining: Analysis, Features, Classification and Retrieval (2019) 35--44.

\bibitem{tripura2023wavelet}
T.~Tripura, A.~Awasthi, S.~Roy, S.~Chakraborty, A wavelet neural operator based elastography for localization and quantification of tumors, Computer Methods and Programs in Biomedicine 232 (2023) 107436.

\bibitem{navaneeth2024waveformer}
N.~Navaneeth, S.~Chakraborty, Waveformer for modeling dynamical systems, Mechanical Systems and Signal Processing 211 (2024) 111253.

\bibitem{wiegreffe2019attention}
S.~Wiegreffe, Y.~Pinter, Attention is not not explanation, arXiv preprint arXiv:1908.04626 (2019).

\bibitem{li2024geometry}
Z.~Li, N.~Kovachki, C.~Choy, B.~Li, J.~Kossaifi, S.~Otta, M.~A. Nabian, M.~Stadler, C.~Hundt, K.~Azizzadenesheli, et~al., Geometry-informed neural operator for large-scale 3d pdes, Advances in Neural Information Processing Systems 36 (2024).

\bibitem{gilmer2017neural}
J.~Gilmer, S.~S. Schoenholz, P.~F. Riley, O.~Vinyals, G.~E. Dahl, Neural message passing for quantum chemistry, in: International conference on machine learning, PMLR, 2017, pp. 1263--1272.

\bibitem{li2020neural}
Z.~Li, N.~Kovachki, K.~Azizzadenesheli, B.~Liu, K.~Bhattacharya, A.~Stuart, A.~Anandkumar, Neural operator: Graph kernel network for partial differential equations, arXiv preprint arXiv:2003.03485 (2020).

\bibitem{nystrom1930praktische}
E.~J. Nystr{\"o}m, {\"U}ber die praktische aufl{\"o}sung von integralgleichungen mit anwendungen auf randwertaufgaben (1930).

\bibitem{daubechies1992ten}
I.~Daubechies, Ten lectures on wavelets, SIAM, 1992.

\bibitem{ncbi_coarctation}
N.~C. for Biotechnology~Information, \href{https://www.ncbi.nlm.nih.gov/books/NBK430913/}{Coarctation of the aorta}, accessed: 2024-12-20 (2024).
\newline\url{https://www.ncbi.nlm.nih.gov/books/NBK430913/}

\end{thebibliography}

\end{document}